\documentclass[letterpaper]{article} % DO NOT CHANGE THIS
\usepackage[draft]{aaai2026}  % DO NOT CHANGE THIS
\usepackage{times}  % DO NOT CHANGE THIS
\usepackage{helvet}  % DO NOT CHANGE THIS
\usepackage{courier}  % DO NOT CHANGE THIS
\usepackage[hyphens]{url}  % DO NOT CHANGE THIS
\usepackage{graphicx} % DO NOT CHANGE THIS
\usepackage{natbib}  % DO NOT CHANGE THIS AND DO NOT ADD ANY OPTIONS TO IT
\usepackage{caption} % DO NOT CHANGE THIS AND DO NOT ADD ANY OPTIONS TO IT
\usepackage{algorithm}
\usepackage{amsmath, amssymb}   % math environments & symbols
\usepackage{enumitem}           % compact lists

\usepackage[dvipsnames]{xcolor}  % 
\usepackage{newfloat}
\usepackage{listings}
\DeclareCaptionStyle{ruled}{labelfont=normalfont,labelsep=colon,strut=off} % DO NOT CHANGE THIS
\floatstyle{ruled}
\newfloat{listing}{tb}{lst}{}
\floatname{listing}{Listing}
\usepackage{algorithm}
\usepackage[noend]{algcompatible}
\usepackage{algpseudocode}
\usepackage{xcolor}
\definecolor{astroblue}{RGB}{0,60,180}
\usepackage{subcaption}   % 提供 subfigure / subtable 环境
\usepackage{booktabs}  % 提供 \toprule \midrule \bottomrule
\usepackage{multirow}  % 提供 \multirow（如果用了）

\usepackage{amsmath,amssymb}  % already allowed by AAAI
\usepackage{amsthm}           % gives theorem/proof environments

\theoremstyle{plain}
\usepackage[table]{xcolor}
\usepackage{colortbl}
\usepackage{booktabs}
\theoremstyle{definition}

\theoremstyle{remark}
\definecolor{rk1}{HTML}{238B45} % Rank-1  最佳，深
\definecolor{rk2}{HTML}{41AB5D}
\definecolor{rk3}{HTML}{74C476}
\definecolor{rk4}{HTML}{A1D99B}
\definecolor{rk5}{HTML}{C7E9C0}
\definecolor{rk6}{HTML}{E5F5E0}
\definecolor{rk7}{HTML}{F7FCF5} % Rank-7  最差，浅

\newcommand{\rankone}{\cellcolor[gray]{0.88}}
\newcommand{\ranksec}{\cellcolor[gray]{0.92}}
\newcommand{\rankthd}{\cellcolor[gray]{0.96}}
\newcommand{\rankfor}{\cellcolor[gray]{1.0}}

\usepackage{enumitem}   % 放在 \documentclass 之后
\title{ASTRO: Adaptive Stitching via Dynamics-Guided Trajectory Rollouts}

\author{
    Hang Yu\textsuperscript{\rm 1},
    Di Zhang\textsuperscript{\rm 1},
    Qiwei Du\textsuperscript{\rm 2},
    Yanping Zhao\textsuperscript{\rm 1},
    Hai Zhang\textsuperscript{\rm 3}\\
    Guang Chen\textsuperscript{\rm 1},
    Eduardo E. Veas\textsuperscript{\rm 4},
    Junqiao Zhao\textsuperscript{\rm 1}\thanks{Corresponding at zhaojunqiao@tongji.edu.cn}
}

\affiliations{
    \textsuperscript{\rm 1}Tongji University, 
    \textsuperscript{\rm 2}University at Buffalo,
    \textsuperscript{\rm 3}The University of Hong Kong,
    \textsuperscript{\rm 4}Graz University of Technology\\
}

\usepackage{bibentry}
\begin{document}

\maketitle

\begin{abstract}

Offline reinforcement learning (RL) enables agents to learn optimal policies from pre-collected datasets. 
However, datasets containing suboptimal and fragmented trajectories present challenges for reward propagation, resulting in inaccurate value estimation and degraded policy performance. 
While trajectory stitching via generative models offers a promising solution, existing augmentation methods frequently produce trajectories that are either confined to the support of the behavior policy or violate the underlying dynamics, thereby limiting their effectiveness for policy improvement. 
We propose ASTRO, a data augmentation framework that generates distributionally novel and dynamics-consistent trajectories for offline RL. 
ASTRO first learns a temporal-distance representation to identify distinct and reachable stitch targets. 
We then employ a dynamics-guided stitch planner that adaptively generates connecting action sequences via Rollout Deviation Feedback, defined as the gap between target state sequence and the actual arrived state sequence by executing predicted actions, to improve trajectory stitching's feasibility and reachability.
This approach facilitates effective augmentation through stitching and ultimately enhances policy learning. 
ASTRO outperforms prior offline RL augmentation methods across various algorithms, achieving notable performance gain on the challenging OGBench suite and demonstrating consistent improvements on standard offline RL benchmarks such as D4RL.

\end{abstract}

% \section{1\quad Introduction}
\section{Introduction}
\begin{figure}[!t]\centering\includegraphics[width=\linewidth]{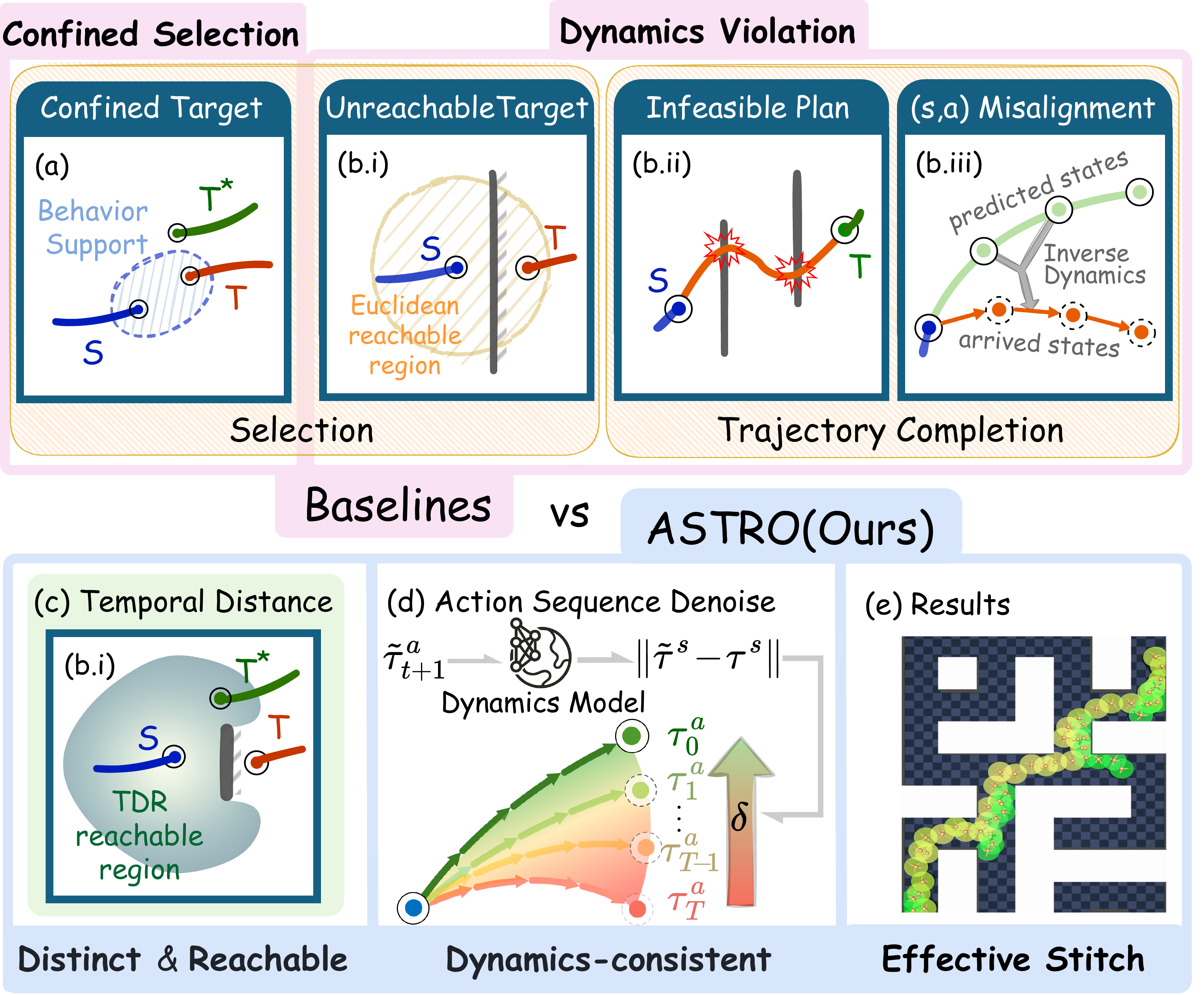}
  \caption{\textbf{Comparison of trajectory stitching approaches}:  
(a) Existing methods suffer from \textit{Confined Target Selection} within behavior policy support; 
(b) \textit{Dynamics Violation} manifests as: i. Infeasible Target Selection via Euclidean metrics (unreachable targets within fixed timesteps behind walls), ii. Infeasible Planning without explicit dynamics modeling, iii. Action-State Misalignment from suboptimal inverse dynamics;
\textbf{ASTRO overcomes these via}: (c) Temporal Distance Representation for distinct, reachable target selection beyond behavior support, and (d) Dynamics-Guided Stitching with action sequence denoising using rollout deviation feedback; 
(e) Resulting in \textit{dynamics-consistent} augmented trajectories that enable effective policy learning.}
  \label{fig:intro}
  \vspace{-0.5cm}
\end{figure}

Offline reinforcement learning (RL) enables agents to acquire decision-making capabilities from pre-collected datasets, thereby avoiding the expense and safety risks associated with direct environment interaction~\citep{levine2020offline,agarwal2020optimistic,td3bc_fujimoto2021,tt_janner2021,morel_kidambi2020}. 
However, in the absence of online exploration, offline RL faces two persistent challenges: distributional shift and value-function overestimation~\citep{levine2020offline,td3bc_fujimoto2021,iql_kostrikov2022}. 
To address these issues, most methods cast offline RL as a constrained optimization problem: maximizing expected returns while restricting the policy to remain within the dataset's state-action distribution~\citep{td3bc_fujimoto2021,iql_kostrikov2022}.

Nevertheless, when datasets consist of suboptimal and temporally fragmented trajectories, reward signals cannot propagate across trajectory boundaries. This impairs value estimation, disrupts long-horizon consistency, and ultimately degrades policy performance.

A promising solution is trajectory stitching, which augments the dataset by synthesizing new trajectories through bridging desirable sub-trajectories~\citep{diffstitch_li2024}. Prior works~\citep{diffstitch_li2024} typically identify source-target stitch pairs by generating candidate states from source trajectories and using Euclidean distance metrics to determine viable targets. Synthetic trajectories are then completed using state-based planners and inverse dynamics models, thereby enriching the data for policy learning.

However, these techniques often underperform in environments with complex dynamics or multi-modal behavior policies~\citep{levine2020offline}. As illustrated in Fig.~\ref{fig:intro}, stitched trajectories suffer from two major issues:
\textbf{(a) Confined Target Selection}:  
Existing approaches constrain stitching to the support of behavior policy by relying on behavior-cloning rollouts to pre-generate candidate targets, limiting novelty and potential policy improvement.
\textbf{(b) Dynamics Violation}:  
This manifests in three aspects:
\textit{i. Infeasible Target Selection}: Euclidean proximity fails to reflect temporal or semantic feasibility in high-dimensional state space. It may select spatially close yet unreachable states within fixed timesteps (e.g., behind obstacles).
\textit{ii. Infeasible Planning}: Previous completion methods entangle policy and dynamics modeling, using state-based planners without explicit dynamics modeling, often producing infeasible plans.
\textit{iii. Action-State Misalignment}: Noisy state prediction and suboptimal inverse dynamics lead to inaccurate actions, whose errors compound over long horizons and cause misalignment between planned state and action sequence.

These limitations underscore our key insight: effective trajectory stitching requires breaking through the constraints of behavior policies and establishing explicit alignment between planning decisions and the environment's underlying dynamics.
We propose ASTRO (\textbf{A}daptive \textbf{S}titching via dynamics-guided \textbf{T}rajectory \textbf{R}oll-\textbf{O}uts), a model-based data augmentation framework designed to generate distributionally novel and dynamics-consistent stitch trajectories for offline RL learning. ASTRO resolves the aforementioned key limitations as follows:
\textbf{(1) Stitch Target Selection in Temporal-Distance-Space}: Instead of relying on pre-generated rollouts and naive distance metrics, ASTRO performs stitch target selection via \textit{Temporal Distance Representation (TDR)}, identifying distinct and reachable sub-trajectories beyond the behavior distribution.
\textbf{(2) Decoupled Planning and Explicit Dynamics Modeling}: Instead of direct state-based completion, ASTRO explicitly separates planning from dynamics modeling, employing a planner to propose action sequences and a long-horizon dynamics model for valid and accurate rollouts.
\textbf{(3) Dynamics-Guided Planning via Rollout Deviation Feedback}: 
ASTRO utilizes dynamics-guided stitch planner that adaptively generates connecting action sequences via \textit{rollout deviation feedback} (i.e. the gap between target states and the actual reached states) to regularize training and enable adaptive inference, thereby ensures stitching feasibility and further improves target reachability.

To our knowledge, ASTRO is the first trajectory stitching method to achieve substantial performance gains on OGBench, a challenging benchmark with complex dynamics and multi-modal behavior policies. While prior methods yield only marginal improvements, ASTRO improves average task performance by 32.7\% (+9.68) across multiple offline RL algorithms. It also provides consistent improvements on standard benchmarks such as D4RL.

% \section{2\quad Related Works}
\section{Related Works}

% \subsection{2.1\quad Offline Reinforcement Learning}
\subsection{Offline Reinforcement Learning}

Offline RL tackles the “distribution-shift dilemma”: maximize return while staying inside the support of a fixed dataset.~\citep{levine2020offline}
Previous works have implemented this high-level objective in diverse ways through
behavioral regularization~\citep{awac_nair2020, td3bc_fujimoto2021, rebrac_tarasov2023},
conservatism~\citep{cql_kumar2020},
in-sample maximization~\citep{iql_kostrikov2022, sql_xu2023, xql_garg2023},
out-of-distribution detection~\citep{MOPO_2020, morel_kidambi2020, edac_an2021, sacrnd_nikulin2023},
dual RL~\citep{optidice_lee2021, dualrl_sikchi2024},
and generative modeling~\citep{dt_chen2021, tt_janner2021, diffuser_janner2022, flowql_park2025, li2024towards_neurips}.
While these methods show promise, they treat trajectories independently, overlooking optimal behavior reconstruction from suboptimal segments. ASTRO addresses this via dynamics-guided trajectory stitching, ensuring both novelty and feasibility.

% \subsection{2.2\quad Trajectory Stitching}
\subsection{Trajectory Stitching}

Recently, many works have explored the trajectory stitching problem given offline data in both implicit and explicit ways. Some methods execute stitchability implicitly during planning, using a Q-function \citep{ssd_kim2024}, condition flow \citep{compdiffuser_luo2025}, dynamics model dreaming \citep{mbrcsl_zhou2023}, or Temporal-Distance based graph \citep{gas_baek2025}.
Another category of solution is based on data augmentation. Recent advances in generative models like Diffusion \cite{song2021sde,ho2020ddpm} have enabled high-quality augmentation. Some works focus on high-reward transition ~\citep{synthER_lu2023}, some directly generate trajectory ~\citep{gta_lee2024,diffstitch_li2024}. To improve generation quality and diversity, suitable for learning policy, ~\citep{rtdiff_yang2025,bitrajdiff_qing2025} explore different generation directions, ~\citep{scots_lee2025} use temporal distance latent space, ~\citep{pgd_jackson2024} performing guidance to narrow policy shift. There are also model-based variants, using dynamics model to perform reachable constrained roll-out ~\citep{MOPO_2020,morel_kidambi2020,combo_yu2021,bats_char2022,leq_park2024,mbts_hepburn2022}. 
Unlike these approaches, ASTRO uniquely combines temporal-distance-based target selection with explicit dynamics modeling and rollout deviation feedback to generate novel yet dynamics-consistent trajectories.

\begin{figure*}[htbp]
    \includegraphics[width=\linewidth]{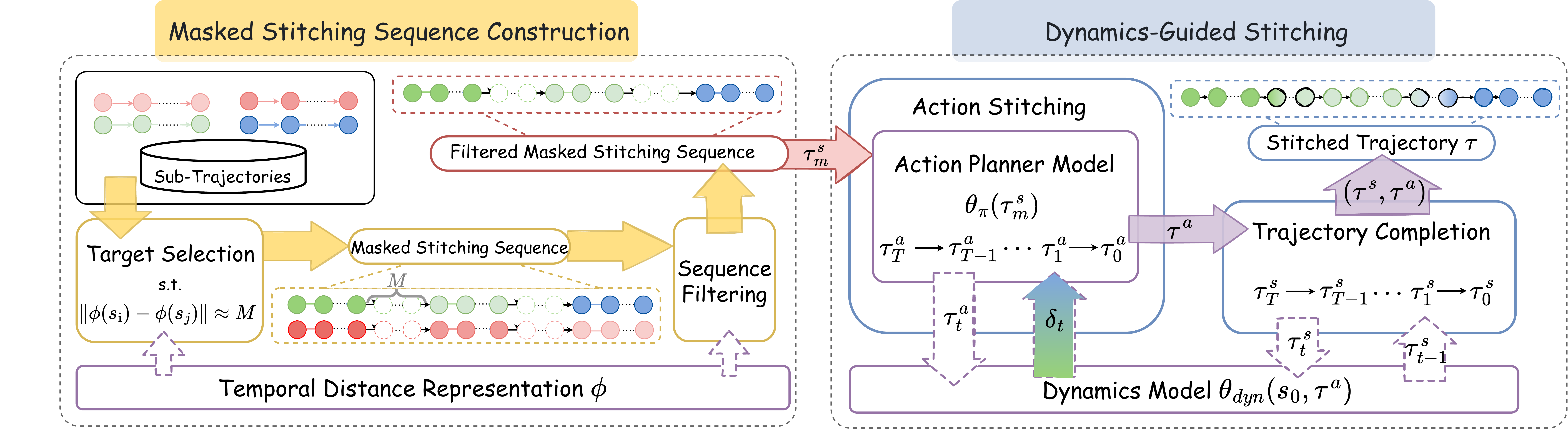}

    \caption {Overview of the \textbf{ASTRO} pipeline: (a) stitch target selection based on Temporal Distance Representation (TDR) for temporal coherence. (b) dynamics-guided trajectory completion using diffusion models with rollout deviation for ensuring feasible, high-quality trajectories.}
    \label{fig:pipeline}
    \vspace{-0.5cm}
  \end{figure*} 
% \section{3\quad Preliminaries}
\section{Preliminaries}
\label{sec:prelim}

% \subsection{3.1\quad Offline Reinforcement Learning}
\subsection{Offline Reinforcement Learning}
\label{sec:prelim_offline_rl}

We consider an infinite–horizon Markov Decision Process (MDP)
$\langle\mathcal{S}, \mathcal{A}, \rho_{0}, p, r, \gamma\rangle$,
where $\mathcal{S}$ and $\mathcal{A}$ denote the state and action spaces,
$\rho_{0}$ is the initial state distribution,
$p$ the transition dynamics,
$r$ the reward function,
and $\gamma \in (0,1)$ the discount factor.
At each timestep $t$, the agent selects an action
$a_t \sim \pi(\cdot \mid s_t)$,
receives reward $r_t$,
and transitions to the next state
$s_{t+1} \sim p(\cdot \mid s_t, a_t)$,
thus forming a trajectory
$\tau = (s_0, a_0, r_0, s_1, \dots)$.
The objective is to learn a policy
$\pi^{\star} = \arg\max_{\pi}
    \mathbb{E}_{\tau \sim \pi}
    \bigl[\sum_{t=0}^{\infty} \gamma^{t} r_t\bigr]$.

In \emph{offline RL}~\citep{levine2020offline}, the agent is given a fixed dataset
$\mathcal{D} = \{\tau_i\}_{i=1}^{N}$
collected by an unknown behavior policy $\pi_{\beta}$ and cannot interact with the environment further.
A learning algorithm is applied to $\mathcal{D}$ and returns a policy
$\pi_{\theta}$.
Its performance is evaluated by:
\begin{equation}
J(\pi_{\theta}) = 
  \mathbb{E}_{\tau \sim \pi_{\theta}}
  \Bigl[\sum_{t=0}^{\infty} \gamma^{t} r_t\Bigr].
\label{eq:offline_return}
\end{equation}
Offline RL algorithms must balance return maximization with staying close to the data distribution to avoid distributional shift and value overestimation~\citep{iql_kostrikov2022, morel_kidambi2020, cql_kumar2020}.

% \subsection{3.2\quad Temporal Distance Representation}
\subsection{Temporal Distance Representation}
\label{sec:prelim_tdr}

When datasets consist of fragmented trajectories, identifying proper \emph{source–target} pairs becomes critical for trajectory stitching.
We adopt the \emph{Temporal Distance Representation} (TDR) proposed by~\citep{hilbert_paper,tldr_paper,cudc_paper,tempdata_paper}, which embeds each state $s$ into a latent space $H$
via a mapping $\psi: \mathcal{S} \to H$ such that
\begin{equation}
d^{\ast}(s, g) = \|\psi(s) - \psi(g)\|_2
\label{eq:td_metric}
\end{equation}
approximates the minimum number of environment steps required to reach $g$ from $s$.

Learning $\psi$ can be formulated as a goal-conditioned value function estimation problem~\citep{hilbert_paper}, optimized with an expectile TD loss over offline triples $(s, s', g)$:
\begin{equation}
\mathcal{L}_{\text{TDR}} =
  \mathbb{E}_{(s, s', g) \sim \mathcal{D}}
  \Bigl[
    l_{\tau}^{2}
    \bigl(-\mathbf{1}\{s \neq g\} + \gamma V(s', g) - V(s, g)\bigr)
  \Bigr],
\end{equation}
where $V(s, g) = -\|\psi(s) - \psi(g)\|_2$ and
$l_{\tau}^{2}(\cdot)$ denotes the expectile regression loss.

By learning temporal consistency across fragmented trajectories, TDR provides a robust distance metric that generalizes beyond behavior policy limitations.

% \subsection{3.3\quad Diffusion Models}
\subsection{Diffusion Models}
\label{sec:prelim_diffusion}

\paragraph{\quad Gaussian diffusion}
A $T$-step diffusion model
corrupts a clean sample $\mathbf x_{0}$
into $\mathbf x_{1},\dots,\mathbf x_{T}$
through
$q(\mathbf x_{t}\!\mid\!\mathbf x_{t-1})
 =\mathcal N(\sqrt{1-\beta_t}\mathbf x_{t-1},\beta_t\mathbf I)$.
The denoiser $\boldsymbol\varepsilon_{\theta}$ is trained \emph{either} to predict the added noise with the standard objective
$\mathbb{E}_{t,\mathbf{x}_{0},\boldsymbol\varepsilon}
       [\lVert\boldsymbol\varepsilon-
              \boldsymbol\varepsilon_{\theta}(\mathbf{x}_{t},t)\rVert_{2}^{2}]
$
or \emph{alternatively} to predict the clean sample, minimising
$\mathbb{E}_{t,\mathbf{x}_{0}}
       [\lVert\mathbf{x}_{0}-\tilde{\mathbf{x}}_{t}\rVert_{2}^{2}]$,
where
\begin{equation}
\label{eq:prelim_clean_t_output}
\displaystyle
  \tilde{\mathbf{x}}_{t}
  =\tilde{\mathbf{x}}_{\theta}(\mathbf{x}_{t},t)
  =\frac{\mathbf{x}_{t}
         -\sqrt{1-\bar\alpha_{t}}\,
          \boldsymbol\varepsilon_{\theta}(\mathbf{x}_{t},t)}
        {\sqrt{\bar\alpha_{t}}}\,
\end{equation}

Following the denoising diffusion framework to predict the clean sample, we first estimate the clean sample $\tilde{\mathbf{x}}_{t}$ from the noisy input $\mathbf{x}_t$ using the model-predicted noise $\boldsymbol{\varepsilon}_\theta(\mathbf{x}_t, t)$ and then obtain the denoised sample at the previous step $t{-}1$ by injecting a small amount of noise back to $\tilde{\mathbf{x}}_t$, following the reverse process of the forward diffusion:
\begin{equation}
\label{eq:xtminus1_from_x0}
\hat{\mathbf{x}}_{t-1}
=
\sqrt{\bar{\alpha}_{t-1}} \, \tilde{\mathbf{x}}_{t} + \sqrt{1 - \bar{\alpha}_{t-1}} \cdot \mathbf{z}, \quad \mathbf{z} \sim \mathcal{N}(0, \mathbf{I}).
\end{equation}

This formulation reuses the predicted clean sample to deterministically guide the generation of $\hat{\mathbf{x}}_{t-1}$, and is widely adopted in DDPM implementations for sampling.

Sampling starts from $\mathbf x_{T}$
and iteratively denoises to $\mathbf x_{0}$.

% \input{sections/2_3_prelim_and_rw}
% \section{4\quad Method}
\section{Method}

We aim to augment offline RL datasets via trajectory stitching, which connects fragmented sub-trajectories into longer, coherent trajectories.
Given a source sub-trajectory
$\tau_{\text{sub}} = (s_i, s_{i+1}, \dots, s_{i+l-1})$,
our goal is to generate a target sub-trajectory $\tau_{\text{sub}}’$ by inserting a masked segment \texttt{MASK} that bridges the two. This masked segment is completed by a generative model, which predicts the missing transitions to ensure dynamic consistency.

However, existing stitching approaches often rely on behavior-cloned rollouts and Euclidean distance metrics to select stitching targets. These heuristics tend to produce temporally incoherent connections, violate environment dynamics, and remain constrained to the behavior policy distribution,thereby limiting their effectiveness.

To overcome these limitations, our method ASTRO selects stitching targets using a learned Temporal Distance Representation (TDR), where latent-space distances approximate temporal distance in environment steps.
Given a source sub-trajectory, ASTRO identifies target states approximately $M$ steps away in TDR space, inserts masked transitions, and filters candidate masked Stitching sequences based on TDR-step consistency.
A diffusion-based planner then completes the masked segment by generating action sequences guided by a long-horizon dynamics model, ensuring feasible and temporally consistent rollouts.

\subsection{ASTRO Stitch Pipeline}
\subsubsection{Masked stitching Sequence Construction}
\label{sec:temporal_space_stitch_picking}

Effective trajectory stitching critically depends on the selection of fragments that are both novel and consistent with environment dynamics. We perform target selection in temporal space using a TDR encoder $\psi: \mathcal{S} \rightarrow \mathbb{R}^d$, which maps states into a latent space where Euclidean distances approximate optimal temporal differences. This process involves two main components: target selection and sequence filtering.

\subsubsection{Target Selection}

To bridge sub-trajectories, we identify a target state $s_{\text{target}}$ approximately $M$ steps away in TDR space from the terminal state $s_{\text{end}}$ of the current sub-trajectory:
\begin{equation}
s_{\text{target}} = \mathop{\arg\min}_{s \in \mathcal{D}} \left| \|\psi(s_{\text{end}}) - \psi(s)\|_2 - M \right|
\end{equation}

We then insert a mask sequence \texttt{MASK} of length $M$ between the source and target sub-trajectories. The resulting masked stitching sequence is structured as:
\begin{equation}
\tau_{\text{m}}^s = (\tau_{\text{sub}}^0, \texttt{MASK}, \tau_{\text{sub}}^1, \texttt{MASK}, \dots)
\end{equation}
By leveraging TDR’s distance approximation $|\psi(s_i) - \psi(s_j)|_2 \approx d^*(s_i, s_j)$ and its ability to encode generalizable temporal reachability, this approach identifies coherent and reachable targets beyond the support of the behavior policy, enabling dynamic-consistent stitching.

\subsubsection{Sequence Filtering}

To further ensure smooth and reliable stitching, we apply TDR-based distance filtering to prune unsuitable stitching sequences. For each candidate stitching sequence, we sample $k$ random state pairs $(s_m, s_n)$ both within and across sub-trajectories, then compute the expected temporal distance bias:
\begin{equation}
\mathbb{E}[\Delta_d] = \mathbb{E}\left[\left|(m-n) - \|\psi(s_m) - \psi(s_n)\|_2 \right|\right]
\end{equation}
Sequences where $\mathbb{E}[\Delta_d] > \Delta_{\text{thresh}}$ are discarded. This enforces local temporal consistency and prevents stitching over structurally inconsistent subsequences (filtering algorithm details in Appendix A).

\subsubsection{Dynamics-Guided Stitching}
\label{sec:dynamics_guided_completion}

\paragraph{Action stitching}

Given a masked stitching sequence \(\tau_{\text{m}}^s\), ASTRO’s stitch planner \(\theta_\pi\) generates an action trajectory \(\tau^a\) via an adaptive denoising process:

\begin{equation}
  \label{eq:self_condition_sample_inference}
  \hat{\tau}_{t-1}^{a}
  \sim
  p_{\theta_{\pi}}
  \left(
        \hat\tau^{a}_{t-1}\middle|\hat\tau^{a}_{t},\tau_{m}^{s}, t, \delta(\tilde\tau_{0}^{a,(t+1)})
  \right).
  \end{equation}

To ensure dynamic consistency, we introduce Rollout Deviation Feedback \(\delta\).
Given a noisy state sequence $\tau^{s}_{t}$ and predicted action sequence $\tau^a$, we compute the deviation between the desired target states and the predicted rollout generated via a learned diffusion dynamics model $\theta_{\text{dyn}}$, which denoises state sequence following:
\begin{equation}
  \label{eq:state_sequence_generation}
  \hat{\tau}_{t-1}^{s} \sim p_{\theta_{\text{dyn}}}\left(\hat{\tau}_{t-1}^{s} \mid \hat{\tau}_{t}^{s}, \tau^a_\text{aug}, s_0, t\right)
\end{equation}

\begin{equation}
\label{eq:extrapolation_error_computation}
\delta(\tau^{a})
  = \bigl\|\tau^{s} - \tilde{\tau}_{\theta_{\text{dyn}}}^{s}(\tau_{t}^{s},s_0, \tau^{a}, t) \bigr\|_2^{2},
\quad t\sim\mathcal U(0,\,T),
\end{equation}

Here, $\tau^{s}$ refers either to the masked stitching sequence $\tau^{s}_{m}$ or a full trajectory $\tau$, depending on context.
This trajectory-level feedback enables iterative refinement of $\tau^a$, guiding the denoising process toward feasible and reachable actions under the environment’s dynamics.

\paragraph{State Sequence Generation} 

Next, we use the dynamics model $\theta_{\text{dyn}}$ to roll out the predicted action sequence and generate a dynamics-consistent state sequence $\tau^s_{\text{aug}}$. The denoising formulation is:

\begin{equation}
  \label{eq:state_sequence_generation}
  \hat{\tau}_{t-1}^{s} \sim p_{\theta_{\text{dyn}}}\left(\hat{\tau}_{t-1}^{s} \mid \hat{\tau}_{t}^{s}, \tau^a_\text{aug}, s_0, t\right)
\end{equation}

By iteratively denoising, the model reconstructs the complete state trajectory $\tau^s_{\text{aug}}$, ensuring temporal coherence aligned with dynamics.

\paragraph{Trajectory Completion}

The final augmented trajectory $\tau_{\text{aug}} = (\tau^s_{\text{aug}}, \tau^a_{\text{aug}})$ is added to the augmentation buffer $\mathcal{D}_{\text{aug}}$. During training, we progressively update the dataset:
$\mathcal{D} \leftarrow \mathcal{D} \cup \mathcal{D}_{\text{aug}}$. 
The augmented dataset is then used to train the policy using standard offline RL algorithms (implementation details in Appendix A).

% \subsection{4.2\quad Model Implementation and Training}
\subsection{Model Implementation and Training}
\label{sec:guided_stitch_planner}

We now present the architecture and training methodology for the dynamics model $\theta_{\text{dyn}}$ and the stitch planner $\theta_\pi$. The planner learns to generate dynamics-consistent action sequences by denoising noisy inputs with guidance from the learned dynamics model.

% \subsubsection{4.2.1\quad Dynamics Diffusion Model}
\subsubsection{Dynamics Diffusion Model}

To provide reliable dynamics feedback and rollouts, we first train a sequence-level dynamics diffusion model $\theta_{\text{dyn}}$ to reconstruct full state trajectories, conditioned on the initial state $s_0$ and the corresponding action sequence $\tau^a$. The model is trained to minimize the diffusion reconstruction loss:
\begin{equation}
  \label{eq:dyn_diffusion_loss}
  \left\{
  \begin{aligned}
  \mathcal{L}_{\text{diff}}(\theta_{\text{dyn}}) &= \mathbb{E}_{t,\tau_s,\tau_a}\left[ \left\| \tau^{s} - \tilde{\tau}_{\theta_{\text{dyn}}}^{s}(\tau_{t}^{s},s_0, \tau^{a}, t) \right\|_2^2 \right] \\
  \tilde{\tau}_{\theta_{\text{dyn}}}^{s} &= \frac{1}{\sqrt{\bar{\alpha}_t}} \left( \tau_{t}^{s} - \sqrt{1 - \bar{\alpha}_t} \, \epsilon_{\theta_{\text{dyn}}}(\tau_{t}^{s}, s_0, \tau^{a}, t) \right)
  \end{aligned}
  \right.
\end{equation}
where $\tau_{t}^{s}$ denotes the state trajectory corrupted by forward noising at diffusion step $t$, $s_0$ is the initial state, $\tau^{a}$ represents the associated action sequence, and $t$ indicates the diffusion step. 

This sequence-level formulation explicitly models long-term dynamics, enabling effective state rollouts without accumulated compounding errors.

% \subsubsection{4.2.2\quad Stitch Planner Training}
\subsubsection{Stitch Planner Training}
The stitch planner $\theta_\pi$ learns to denoise a noisy action sequence and generate a goal-directed action plan aligned with the learned dynamics model. Training is supervised via trajectory reconstruction and rollout deviation minimization.

\paragraph{Adaptive Reconstruction Loss}
\label{sec:self_conditioned_action_reconstruction_loss}
We leverage the Rollout Deviation Feedback $\tau^s_{m}$ from Sec.~\ref{sec:dynamics_guided_completion}) to guide denoising. The adaptive self-correction loss is formulated as:

\begin{equation}
\label{eq:self_conditioned_action_reconstruction_loss}
\begin{aligned}
\mathcal{L}_{\text{sc}}(\theta_{\pi})
&\!=\!
  \mathbb{E}_{t,\tau_0}\Bigl[
  \bigl\|
      \tau^{a}
      -\tilde{\tau}_{\theta_{\pi}}^{a}
        \bigl(
          \tau_{t}^{a},\,
          \tau^{s}_{m},\,
          t,\,
          \operatorname{sg}\bigl[\delta(\tau^{a})\bigr]
        \bigr)
    \bigr\|_2^{2}
  \\[2pt]
  &\!+
    \bigl\|
      \tau^{a}
      -\tilde{\tau}_{\theta_{\pi}}^{a}
        \bigl(
          \operatorname{sg}\!\bigl[\hat{\tau}_{t-1}^{a}\bigr],\,
          \tau^{s}_m,\,
          t - 1,\,
          \operatorname{sg}\bigl[\delta(\tilde{\tau}_{t}^{a})\bigr]
        \bigr)
    \bigr\|_2^{2}
  \Bigr]
\end{aligned}
\end{equation}
where \(\hat{\tau}_{t-1}^{a}\) and \(\tilde{\tau}_{t}^{a}\) follows Eq.\ref{eq:xtminus1_from_x0},
\(\operatorname{sg}[\cdot]\) denotes the stop-gradient operator.
The first term encourages accurate prediction despite model error, while the second facilitates recursive correction between denoising steps.
This loss encourages robust denoising by incorporating trajectory-level deviation signals, helping the planner iteratively correct toward feasible action sequences.

\paragraph{Deviation Regularization} 
\label{sec:extrapolation_error_regularization_loss}

To further promote feasible and reliable stitching, we penalize generated actions whose predicted rollouts exhibit larger deviation from the target states than those produced by the ground-truth actions:

\begin{equation}
\label{eq:extrapolation_error_regularization_loss}
\mathcal{L}{\text{reg}}(\theta\pi) = \mathbb{E}{t,\tau_0} \left[ \left( \delta(\tilde{\tau}_{0}^{a,(t)}) - \delta(\tau^{a}) \right)_+ \right],
\end{equation}
where $(x)_+ = \max(0, x)$ is the ReLU operator. This term penalizes the stitch planner only when generated plans degrade reachability relative to expert actions.
This regularization, applied within the model’s confidence region, improves stitching reliability and enhances downstream task performance.

% \paragraph{Joint Training Objective with Rollout Deviation Feedback.} 
\paragraph{Joint Training Objective} 
\label{sec:joint_loss}
% Combining these terms, the final joint training objective is:

The final objective for training the stitch planner is a weighted combination of the self-correction and regularization losses:

\begin{equation}
\label{eq:joint_loss}
\mathcal{L}(\theta_\pi) = \mathcal{L}_{\text{sc}}(\theta_\pi) + \alpha\mathcal{L}_{\text{reg}}(\theta_\pi),
\end{equation}
where $\alpha$ is a hyperparameter that controls the trade-off between trajectory accuracy and dynamics-aligned feasibility.

% \section{5\quad Experiments}
\section{Experiments}

\begin{table*}[t]
\centering
\caption{\small Comparison of ASTRO against baselines (DiffStitch, SynthER) across OGBench and D4RL benchmarks, evaluated with offline RL algorithms IQL and FQL. Results highlight ASTRO's average task performance improvements in various locomotion and manipulation tasks.}
\setlength{\tabcolsep}{2.8pt}\renewcommand{\arraystretch}{1.05}
\begin{tabular}{clcccc|cccc}
\toprule
\multicolumn{2}{c}{\multirow{2}{*}{\large \textbf{Task}}} & \multicolumn{4}{c}{IQL} & \multicolumn{4}{c}{FQL}\\
\cmidrule(lr){3-6}\cmidrule(lr){7-10}
 & & Original & ASTRO & DiffStitch & SynthER & Original & ASTRO & DiffStitch & SynthER \\
\midrule
\multirow{6}{*}{\shortstack{OGBench\\Maze\\Stitch}} 
    & ant-large-v0 & \rankfor 26.2 & \rankone \textbf{ 51.7} & \ranksec 35.0 & \rankthd 31.1 & \ranksec 29.2 & \rankone \textbf{ 57.3} & \rankthd 33.1 & \rankfor 28.7 \\
    & ant-giant-v0 & \rankfor 0 & \rankfor 0 & \rankfor 0 & \rankfor 0 & \ranksec 5.3 & \rankone \textbf{ 10.4} & \rankfor 3.3 & \rankthd 4.2 \\
    & humanoid-medium-v0 & \rankthd 29.7 & \rankone \textbf{ 31.4} & \rankfor 28.3 & \ranksec 31.2 & \rankthd 17.5 & \rankone \textbf{ 30.0} & \ranksec 22.6 & \rankfor 15.9 \\
    & humanoid-large-v0 & \ranksec 2.4 & \rankone \textbf{ 12.6} & \rankthd 2.2 & \rankfor 1.4 & \rankfor 2.7 & \rankone \textbf{ 3.5} & \rankthd 2.9 & \ranksec 3.1 \\
    & antsoccer-arena-v0 & \rankfor 3.7 & \rankone \textbf{ 14.6} & \ranksec 3.9 & \rankthd 3.8 & \rankthd 22.4 & \rankone \textbf{ 49.3} & \ranksec 28.5 & \rankfor 25.7 \\
\cmidrule{2-10}

    & \textbf{maze-avg} & \rankfor 12.40 & \rankone \textbf{ 22.06} & \ranksec 13.88 & \rankthd 13.50 & \rankthd 15.42 & \rankone \textbf{ 30.10} & \ranksec 18.08 & \rankfor 15.52 \\
\midrule
\multirow{4}{*}{\shortstack{OGBench\\Manipulation\\Play}}
    & scene-v0 & \rankthd 31.7 & \rankone \textbf{ 40.6} & \ranksec 32.1 & \rankfor 27.5 & \rankthd 94.3 & \rankone \textbf{ 97.0} & \rankfor 91.4 & \ranksec 92.0 \\
    & cube-single-v0 & \rankthd 81.5 & \rankone \textbf{ 89.2} & \rankfor 79.0 & \ranksec 82.4 & \rankthd 88.0 & \rankone \textbf{ 92.9} & \ranksec 89.6 & \rankfor 87.3 \\
    & cube-double-v0 & \rankthd 2.4 & \ranksec 2.5 & \rankone \textbf{ 2.6} & \rankfor 0.5 & \rankthd 36.5 & \rankone \textbf{ 45.4} & \ranksec 40.1 & \rankfor 41.1 \\
\cmidrule{2-10}
    & \textbf{manipulation-avg} & \ranksec 38.53 & \rankone \textbf{ 44.10} & \rankthd 37.9 & \rankfor 36.8 & \rankthd 72.93 & \rankone \textbf{ 78.43} & \ranksec 73.7 & \rankfor 73.5 \\
\midrule
\multicolumn{2}{l}{D4RL \textbf{avg} (6 ant-mazes)} & \rankfor 57.3 & \rankone \textbf{ 70.4} & \ranksec 65.3 & \rankthd 63.5 & \rankthd 78.2 & \rankone \textbf{ 88.6} & \ranksec 85.0 & \rankfor 79.0 \\
\midrule
\multicolumn{2}{l}{\textbf{Total avg}} & \rankfor 36.08 & \rankone \textbf{ 45.52} & \ranksec 39.03 & \rankthd 37.93 & \rankthd 55.52 & \rankone \textbf{ 65.71} & \ranksec 58.93 & \rankfor 56.01 \\
\bottomrule
\end{tabular}

\label{tab:astro}
\end{table*}

\begin{table*}[htbp]

    \centering
    \caption{\small \textbf{Ablation study} on ASTRO’s stitching mechanism using FQL across four challenging OGBench locomotion tasks. Column labels denote, \textit{Ori}: original dataset performance, ASTRO: our full method, \textit{Rand}: random target selection, Euc: Behavioral Pre-generation + Euclidean-distance-based target selection, \textit{MB}: model-based rollout without guidance, \textit{SI}: state planner with inverse dynamics. ASTRO consistently outperforms all ablations, demonstrating the importance of both Temporal-distance-space target selection and dynamics-guided trajectory stitching.}
% \footnotesize
\setlength{\tabcolsep}{5pt}
\renewcommand{\arraystretch}{1.1}
\begin{tabular}{lcccccc}
\toprule
\textbf{Task} & Ori & \textbf{ASTRO} & w/ Rand & w/ Pre+Euc & w/ MB & w/ SI \\
\midrule
AntMaze-Large-v0       & 29.2  & \textbf{57.3} & 36.2 & 41.3 & 46.5 & 35.9 \\
AntMaze-Giant-v0       &  5.3  & \textbf{10.4} &  7.2 &  4.4 &  6.3 &  8.1 \\
HumanoidMaze-Medium-v0 & 17.5  & \textbf{30.0} & 23.9 & 21.7 & 28.3 & 11.5 \\
AntSoccer-Arena-v0     & 22.4  & \textbf{49.3} & 30.9 & 36.0 & 46.1 & 33.3 \\
\midrule
\textbf{Average}       & 18.60 & \textbf{36.75} & 24.55 & 25.85 & 31.80 & 22.20 \\
\bottomrule
\end{tabular}
\vspace{-0.3cm}
\label{tab:astro_ablation}
\end{table*}

% \subsection{5.1\quad Experimental Setup}
\subsection{Experimental Setup}
\label{sec:exp_setup}

\subsubsection{Benchmarks}

% \john{add citations:}
We primarily evaluate ASTRO on OGBench~\citep{park2024ogbench}, a challenging benchmark consisting of robotic locomotion and manipulation tasks characterized by sparse goal-achievement rewards. We select the reward-based, single-task variants that are compatible with standard offline RL algorithms, specifically including three manipulation and five locomotion tasks.

For locomotion tasks, we utilize the \texttt{stitch} variant datasets, which consist of short, fragmented trajectory segments that require effective stitching to learn coherent long-horizon behavior.

Additionally, we evaluate on six widely used AntMaze tasks from the D4RL benchmark~\citep{fu2020d4rl}, allowing broader comparison against standard baselines. The environment details are illustrated in Appendix B.

\subsubsection{Baselines}  

We compare ASTRO against two state-of-the-art trajectory augmentation methods: \textit{(1)} DiffStitch~\citep{diffstitch_li2024}: A trajectory stitching approach that generates new rollouts via start–goal conditioned diffusion models. \textit{(2)} SynthER~\citep{synthER_lu2023}: A reward-guided diffusion method that augments the replay buffer by synthesizing high-reward transitions.
All methods are evaluated under two popular offline RL algorithms:(1) IQL~\citep{iql_kostrikov2022}: A conservative one-step algorithm based on advantage-weighted behavior cloning.
(2) FQL~\citep{flowql_park2025}: A more expressive algorithm that uses flow-matching-based action sampling.
To ensure fair comparisons, we use identical training protocols, network architectures, and diffusion backbones across all methods (see Appendix B for details).

\subsubsection{Evaluation Protocol}  

Agents are trained for a fixed number of gradient steps, and we report performance using the final checkpoint (rather than selecting the best), to avoid early-stopping bias.

Each reported result represents the mean standard deviation over eight random seeds per task. Complete Evaluation details are provided in Appendix B.

\subsection{Results and Analysis}
\label{sec:exp_results}

We systematically evaluate ASTRO across four key research questions that assess its (1) overall performance in offline RL, (2) underlying improvement mechanisms, (3) reliability of temporal-distance-based target selection, and (4) quality of trajectories generated via dynamics-guided completion. Our results demonstrate how ASTRO’s components synergistically improve trajectory augmentation in offline RL.

\subsubsection{Q1: How significantly does ASTRO enhance offline RL performance?}
\label{sec:perf}

Table~\ref{tab:astro} shows that ASTRO consistently achieves substantial performance gains across locomotion and manipulation tasks with dense or sparse reward, significantly outperforming existing methods. On average, ASTRO boosts scores by \textbf{+26.2\%} under IQL (36.08 $\rightarrow$ 45.52) and \textbf{+18.4\%} under FQL (55.52 $\rightarrow$ 65.71).

ASTRO excels across distinct performance regimes:
\begin{itemize}
  \item \textbf{Moderate-performing tasks} (baseline scores between 20 and 80 on the original dataset): ASTRO achieves substantial improvements, increasing average scores by +15.83 (IQL) and +18.57 (FQL), significantly outperforming DiffStitch (+5.73 IQL, +5.10 FQL) and SynthER (+2.30 IQL, +2.07 FQL).
  
  \item \textbf{Low-return scenarios} (baseline scores below 20 on the original dataset): ASTRO demonstrates strong performance recovery, improving IQL from 3.05 to 13.60 and FQL from 11.40 to 20.20. In comparison, DiffStitch yields marginal gains (IQL: 3.05 $\rightarrow$ 3.05, FQL: 11.40 $\rightarrow$ 12.95), while SynthER leads to performance degradation (IQL: 3.05 $\rightarrow$ 2.60, FQL: 11.40 $\rightarrow$ 10.05).
  
  \item \textbf{High-performing tasks} (baseline scores above 80 on the original dataset): Even in strong-performing regimes, ASTRO achieves further gains of +7.70 (IQL) and +3.80 (FQL), surpassing DiffStitch (--2.50 IQL, --0.65 FQL) and SynthER (+0.90 IQL, --1.50 FQL).
\end{itemize}

% }

Importantly, ASTRO consistently maintains or improves performance across all environments, unlike DiffStitch and SynthER, which occasionally degrade results. This highlights ASTRO’s robustness and reliability in trajectory augmentation.

\subsubsection{Q2: Why does ASTRO work?}
\label{sec:mechanism}

\begin{figure}[h]
  \vspace{-4mm}
  \centering
  \begin{subfigure}[t]{0.32\linewidth}
    \centering
    % TODO: replace with navigate_heatmap.pdf
    \includegraphics[width=\linewidth]{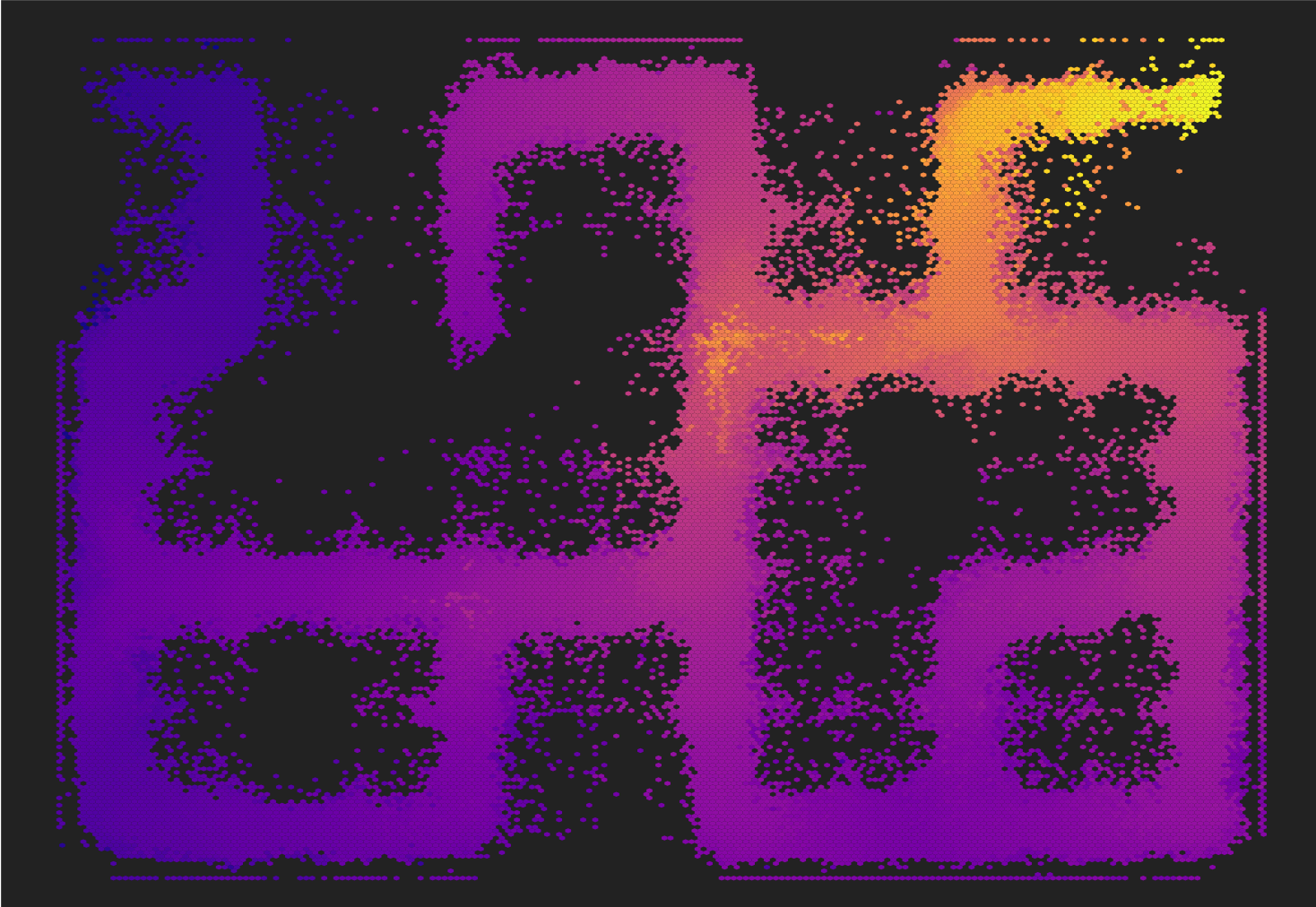}
    \caption{ori}
  \end{subfigure}
  \hfill
  \begin{subfigure}[t]{0.32\linewidth}
    \centering
    % TODO: replace with stitch_heatmap.pdf
    \includegraphics[width=\linewidth]{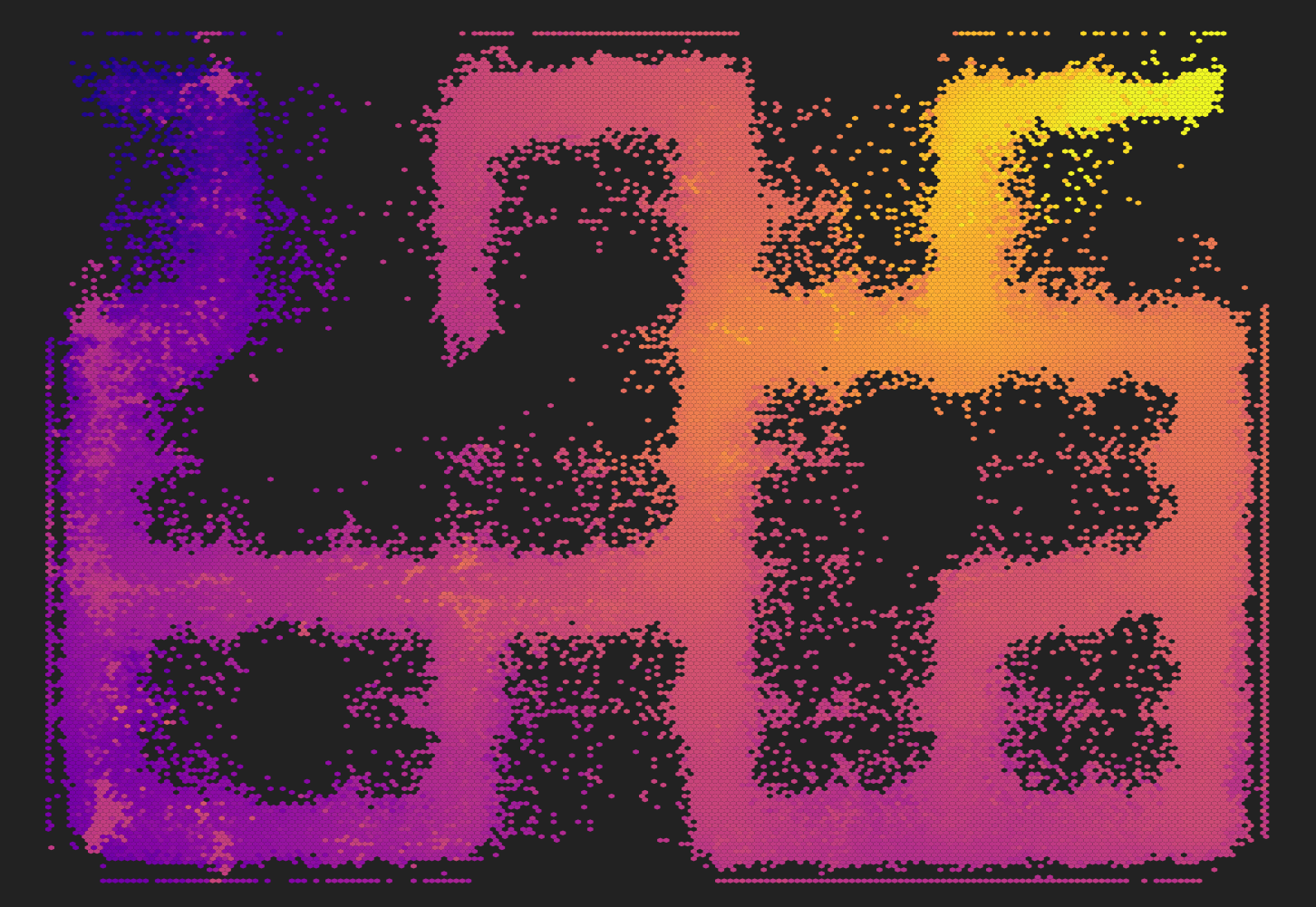}
    \caption{DiffStitch}
  \end{subfigure}
  \hfill
  \begin{subfigure}[t]{0.32\linewidth}
    \centering
    % TODO: replace with astro_heatmap.pdf
    \includegraphics[width=\linewidth]{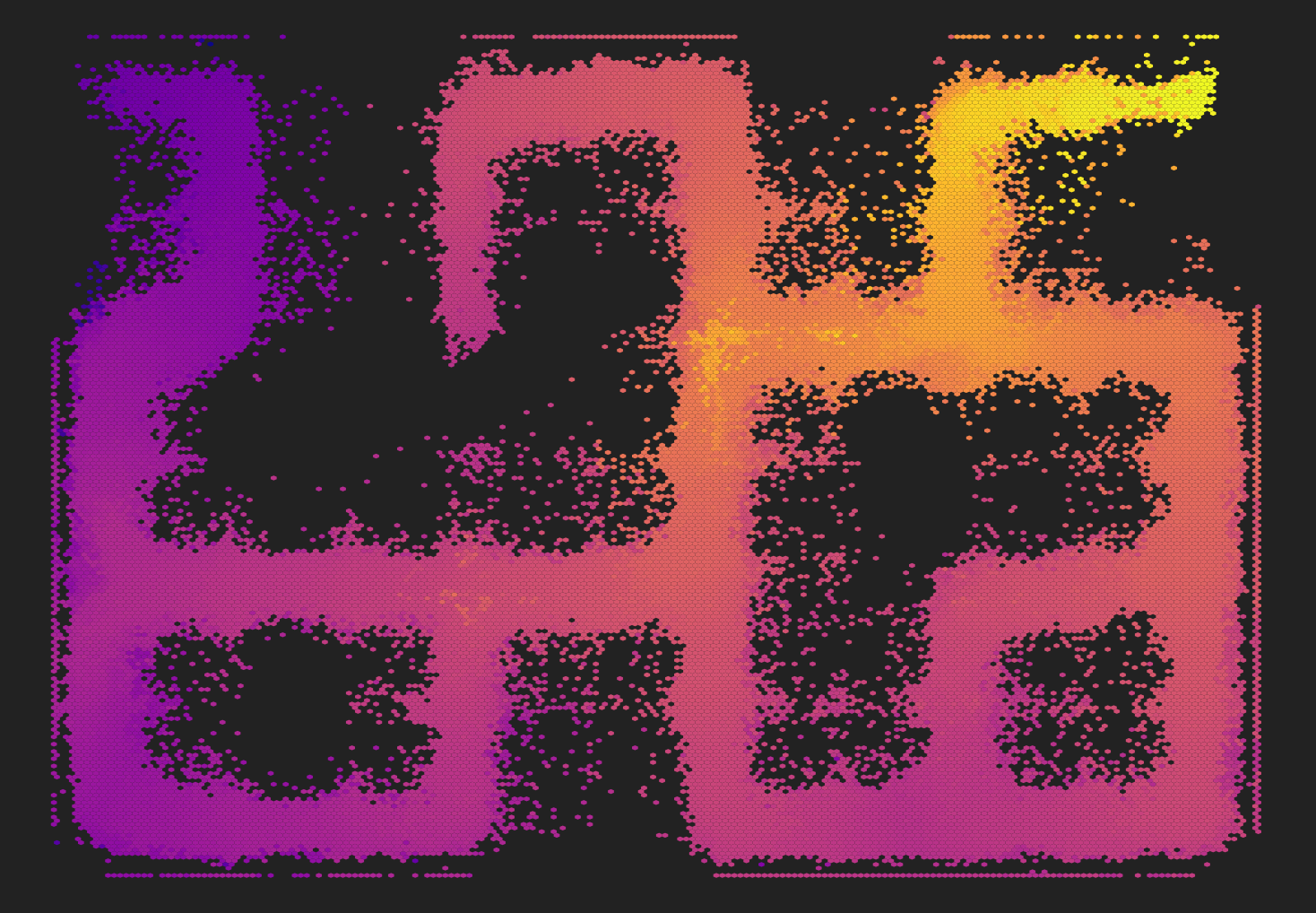}
    \caption{ASTRO}
  \end{subfigure}
\caption{Value-function heatmaps on \texttt{antmaze-large}. Warmer colors indicate higher Q-values; the goal is located in the upper-right corner. ASTRO facilitates effective reward propagation, yielding substantial improvements in \(Q_{\text{mean}}\): \(+16.59\) (from $-111.17$ to $-94.58$) for IQL and \(+7.06\) (from $-87.55$ to $-80.49$) for FQL. In contrast, DiffStitch yields only marginal gains of \(+3.82\) and \(+0.43\), respectively.}
  \vspace{-0.4cm}
  \label{fig:heatmaps}
\end{figure} 

We use the \texttt{antmaze-large} environment from OGBench as a case study to investigate the mechanisms driving ASTRO’s improvements.

In sparse-reward environments with complex dynamics, high-value signals are often concentrated near goal regions, making it difficult for value functions to propagate effectively. ASTRO mitigates this by injecting dynamics-consistent rollouts that expand the set of reachable, high-reward states, thereby promoting more effective Q-value propagation.
As shown in Figure~\ref{fig:heatmaps}, ASTRO significantly improves the distribution of Q-values across the state space, resulting in higher average Q-values and enhanced downstream policy performance. Notably, ASTRO achieves a +9.71 increase in \(Q_{mean}\) compared to baseline methods.

\subsubsection{Q3: How critical is Temporal distance Space for stitch target selection?}
\label{sec:td_space_reliability}

% \begin{figure}[h]
%   \centering
%   % Subfigure (a): Angular deviation visualization
%   \begin{subfigure}[t]{0.32\linewidth}
%     \centering
%     % \includegraphics[width=\linewidth]{fig/angular_deviation}
%     \caption{Angular deviation ($|\Delta\theta|$) comparison}
%     \label{fig:angular_deviation}
%   \end{subfigure}
%   \hfill
  
%   % Subfigure (b): Dynamics violations
%   \begin{subfigure}[t]{0.32\linewidth}
%     \centering
%     % \includegraphics[width=\linewidth]{fig/dynamics_violations}
%     \caption{Dynamics violations}
%     \label{fig:violations}
%   \end{subfigure}
%   \hfill
%   \begin{subfigure}[t]{0.32\linewidth}
%     \centering
%     % \includegraphics[width=\linewidth]{fig/behavior_distribution}
%     \caption{Behavior support breakout}
%     \label{fig:behavior_distribution}
%   \end{subfigure}
%   % % Subfigure (d): Stitch coverage
%   % \begin{subfigure}[t]{0.49\linewidth}
%   %   \centering
%   %   % \includegraphics[width=\linewidth]{fig/stitch_coverage}
%   %   \caption{Stitch coverage comparison}
%   %   \label{fig:stitch_coverage}
%   % \end{subfigure}
  
%   \caption{Analysis of Temporal-Distance-space selection in \texttt{antmaze-large}: 
%   (a)~Temporal-Distance-space selection maintains low angular deviation for smooth paths; 
%   (b)~Temporal-Distance-space selection Avoids infeasible target for dynamics violations; 
%   (c)~Explores beyond behavior distribution boundaries}
%   \label{fig:td_metric_fig}
% \end{figure} 

\begin{figure}[htbp]
    \includegraphics[width=\linewidth]{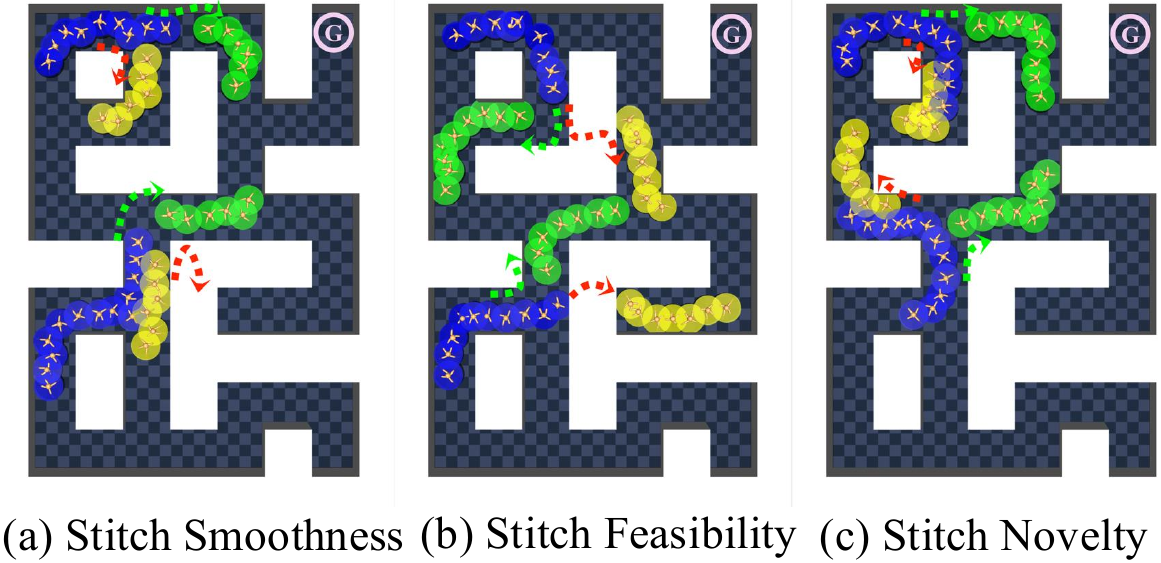}

\caption{Analysis of Temporal-Distance-space selection in \texttt{antmaze-large}: 
  (a)~Temporal-Distance-space selection maintains low angular deviation for smooth paths; 
  (b)~Temporal-Distance-space selection Avoids infeasible target for dynamics violations; 
  (c)~Explores beyond behavior distribution boundaries}
    \label{fig:td_metric_fig}
    \vspace{-0.4cm}
  \end{figure}
\begin{table}[h]
  \centering
  \caption{\small Quantitative analysis of geometric properties for selected stitch source-target pairs in \texttt{antmaze-large} environment, comparing TDR target selection against baseline methods, highlighting improvements in angular deviation and trajectory curvature.}
  \label{tab:td_geom}
  \small
  \setlength{\tabcolsep}{4.8pt}
  \begin{tabular}{lccc}
    \toprule
    \textbf{Selection Method} & 
    $|\Delta\theta|$ ↓&
    Curvature ↑ \\
    \midrule
    Pre-Gen+Eucli & 3.147 $\pm$ 2.621 & 0.697 $\pm$ 0.325 \\
    original traj & 2.089 $\pm$ 1.455 & 0.764 $\pm$ 0.153 \\
    \textbf{TD-Space} & \textbf{1.253 $\pm$ 0.451} & \textbf{0.934 $\pm$ 0.079} \\
    \bottomrule
  \end{tabular}
  \vspace{-0.5cm}
\end{table} 

We analyze the importance of Temporal Distance (TD) space in target selection as follows using \texttt{antmaze-large} as a representative case.

\paragraph{Performance Impact} 
As illustrated in Table~\ref{tab:astro_ablation} Replacing TD-space with alternative selection strategies leads to significant degradation. 
Using Euclidean distance, FQL drops from \(36.75 \rightarrow 25.85\; (-10.9)\). 
Uniform random selection performs even worse, reducing FQL to \(24.55\; (-12.2)\).

\paragraph{Robust Target selection} 

TD space captures temporal coherence, improving stitching quality. 
In contrast, Euclidean matching often yields trajectories with: 
(1) high angular deviation ($|\Delta\theta|$, average direction change between segments) and 
(2) low curvature (inverse turning radius, indicating sharper turns). 
These result in non-smooth paths that violate dynamics constraints (Table~\ref{tab:td_geom}, Fig.\ref{fig:td_metric_fig}(a); calculation details in Appendix).
It may also select unreachable goals, such as those behind walls (Fig.\ref{fig:td_metric_fig}(b))(detailed metric calculation in Appendix C).

\paragraph{Distributional Generalization}

TD-based selection enables ASTRO to go beyond the behavior policy, choosing targets outside the support of the original dataset (Fig.~\ref{fig:td_metric_fig}). This overcomes the limitations of pre-generated, behavior-cloned rollouts and facilitates more diverse and effective stitching.

\subsubsection{Q4: Does dynamics-guided Stitching enhance trajectory quality?}
\label{sec:dyna_guidance}

% \begin{figure}[h]
%   \centering
%   \begin{subfigure}[t]{0.32\linewidth}
%     \centering
%     % \includegraphics[width=\linewidth]{fig/state_inv_dyn}
%     \caption{State planner + inverse dynamics}
%     \label{fig:state_inv_dyn}
%   \end{subfigure}\hfill
%   \begin{subfigure}[t]{0.32\linewidth}
%     \centering
%     % \includegraphics[width=\linewidth]{fig/sequence_model}
%     \caption{Sequence model-based method}
%     \label{fig:sequence_model}
%   \end{subfigure}\hfill
%   \begin{subfigure}[t]{0.32\linewidth}
%     \centering
%     % \includegraphics[width=\linewidth]{fig/astro_adaptive}
%     \caption{ASTRO: Adaptive guidance}
%     \label{fig:astro_adaptive}
%   \end{subfigure}
%   \vspace{-0.6em}
%   \caption{Trajectory completion for identical source-target pair: 
% (a)~State planner + inverse dynamics frequently produces infeasible, unreachable plans with amplified action error; 
% (b)~Sequence model-based method fails to reach the goal; 
% (c)~ASTRO generates reasonable trajectories that successfully reach the target.}
%   \label{fig:dyna_vis}
% \end{figure} 

\begin{figure}[htbp]
    \includegraphics[width=\linewidth]{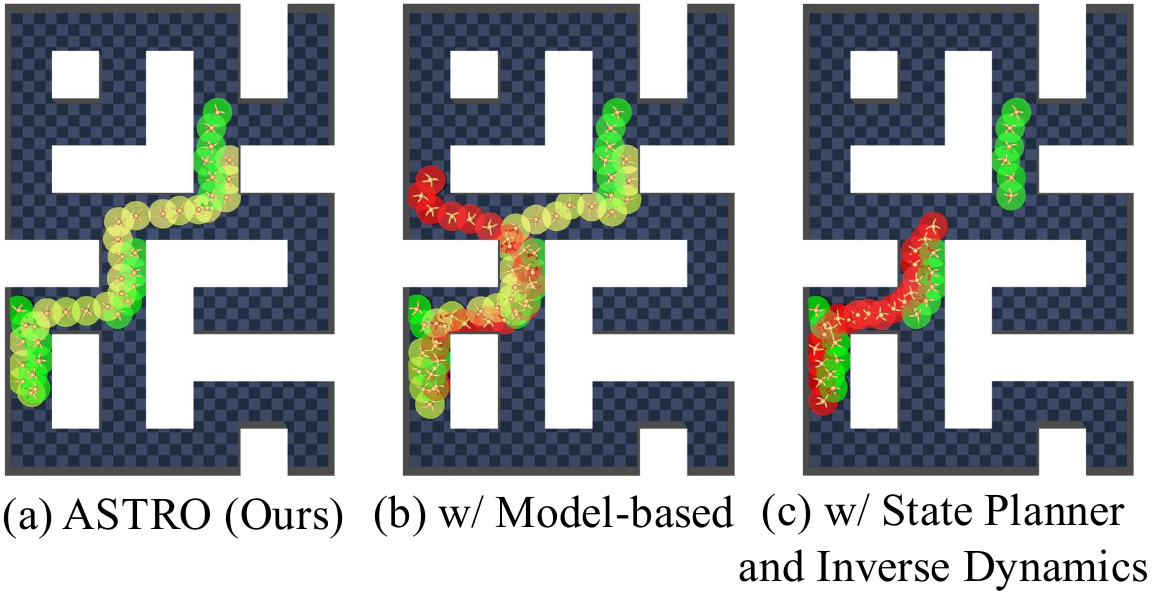}
%   \caption{Trajectory completion for identical source-target pair: 
% (a)~ASTRO generates reasonable trajectories that successfully reach the target.
% (b)~State planner + inverse dynamics frequently produces infeasible, unreachable plans with amplified action error; 
% (c)~Sequence model-based method fails to reach the goal; 
% }
\caption{Completion from different generation pipelines.}
\vspace{-0.1cm}
  \label{fig:dyna_vis}
  \end{figure}
\begin{table}[h]
  \centering
  \caption{\small Evaluation of trajectory completion quality on the \texttt{antmaze-large} task, comparing full ASTRO’s dynamics-guided completion against state-based planners (SI), sequence-model methods (MB), and some ASTRO variants, assessed by action and state prediction errors as well as dynamics violation frequency.}
  \label{tab:dyna_mse}
  \small
  \setlength{\tabcolsep}{4pt}
  \begin{tabular}{lcccc}
    \toprule
    \textbf{Method} & $\tau^{a}$ MSE & $\tau^{s}$ MSE & $\tau^{s}_m$ MSE & Dyn\_Violation \\
    \midrule
    SI & 0.226 & 0.954 & 0.695 & 17.4 \\
    MB & 0.141 & 0.782 & 0.452 & 12.3 \\
    ASTRO (w/ $L_{reg}$) & 0.138 & 0.724 & 0.391 & 9.2 \\
    ASTRO (w/ $L_{sc}$) & 0.129 & 0.723 & 0.402 & 8.5 \\
    \textbf{ASTRO} & \textbf{0.103} & \textbf{0.657} & \textbf{0.351} & \textbf{5.3} \\
    \bottomrule
  \end{tabular}
  \vspace{-0.5cm}
\end{table}

\medskip
\noindent

\paragraph{Feasible Completions}
Table~\ref{tab:dyna_mse} shows that the sequence model alone reduces action MSE by 37.6\% and dynamics violations by 29.3\% compared to inverse dynamics-based stitching. ASTRO further improves on this, reducing action MSE to 0.103 and violations to 5.3\% via training regularization and inference-time self-conditioning.

\paragraph{High Target Reachability}
We assess target reachability by the mean squared error (MSE) between reached states and target states. As shown in Table~\ref{tab:dyna_mse}, ASTRO consistently achieves lower deviation (0.35 MSE) compared to both model-based (0.45 MSE) and inverse dynamics-based stitching (0.70 MSE), which suffer from greater misalignment due to inaccurate action prediction. 
This confirms that ASTRO not only ensures feasibility but also enhances target reachability through its dynamics-guided completion(detailed metric calculation in Appendix C).

\paragraph{Visual Evidence}
Visual evidence in Figure~\ref{fig:dyna_vis} demonstrates ASTRO's dynamics-guided stitching advantages:
(a) ASTRO successfully completes trajectories by leveraging its diffusion-based planner with rollout deviation feedback, ensuring dynamic consistency;
(b) The sequence model-based method fails to reach targets due to lack of dynamics-aware refinement;
(c) The inverse dynamics planner produces state-action misalignment and infeasible plans without explicit dynamics modeling.
These results validate how ASTRO's dynamics-guided completion enables feasible, high-reachability stitching.

% \section{5\quad Conclusion}
\section{Conclusion}

We presented ASTRO, an adaptive trajectory stitching framework for offline reinforcement learning that addresses key limitations of existing augmentation methods. By leveraging a Temporal Distance Representation for  temporally coherent target selection, and employing a dynamics-guided diffusion planner with rollout deviation feedback, ASTRO generates trajectories that are both novel and feasible. This enables more effective value propagation across fragmented sub-trajectories, improving long-horizon policy learning.

\newpage
\newpage

\newpage

\newpage


\bibliography{aaai2026}

@article{awac_nair2020,
  title={Awac: Accelerating online reinforcement learning with offline datasets},
  author={Nair, Ashvin and Gupta, Abhishek and Dalal, Murtaza and Levine, Sergey},
  journal={arXiv preprint arXiv:2006.09359},
  year={2020}
}

@article{td3bc_fujimoto2021,
  title={A minimalist approach to offline reinforcement learning},
  author={Fujimoto, Scott and Gu, Shixiang Shane},
  journal={Advances in neural information processing systems},
  volume={34},
  pages={20132--20145},
  year={2021}
}

@article{rebrac_tarasov2023,
  title={Revisiting the minimalist approach to offline reinforcement learning},
  author={Tarasov, Denis and Kurenkov, Vladislav and Nikulin, Alexander and Kolesnikov, Sergey},
  journal={Advances in Neural Information Processing Systems},
  volume={36},
  pages={11592--11620},
  year={2023}
}

@article{cql_kumar2020,
  title={Conservative q-learning for offline reinforcement learning},
  author={Kumar, Aviral and Zhou, Aurick and Tucker, George and Levine, Sergey},
  journal={Advances in neural information processing systems},
  volume={33},
  pages={1179--1191},
  year={2020}
}

@article{iql_kostrikov2022,
  title={Offline reinforcement learning with implicit q-learning},
  author={Kostrikov, Ilya and Nair, Ashvin and Levine, Sergey},
  journal={arXiv preprint arXiv:2110.06169},
  year={2021}
}

@article{sql_xu2023,
  title={Offline rl with no ood actions: In-sample learning via implicit value regularization},
  author={Xu, Haoran and Jiang, Li and Li, Jianxiong and Yang, Zhuoran and Wang, Zhaoran and Chan, Victor Wai Kin and Zhan, Xianyuan},
  journal={arXiv preprint arXiv:2303.15810},
  year={2023}
}

@article{xql_garg2023,
  title={Extreme q-learning: Maxent rl without entropy},
  author={Garg, Divyansh and Hejna, Joey and Geist, Matthieu and Ermon, Stefano},
  journal={arXiv preprint arXiv:2301.02328},
  year={2023}
}

@article{edac_an2021,
  title={Uncertainty-based offline reinforcement learning with diversified q-ensemble},
  author={An, Gaon and Moon, Seungyong and Kim, Jang-Hyun and Song, Hyun Oh},
  journal={Advances in neural information processing systems},
  volume={34},
  pages={7436--7447},
  year={2021}
}

@inproceedings{sacrnd_nikulin2023,
  title={Anti-exploration by random network distillation},
  author={Nikulin, Alexander and Kurenkov, Vladislav and Tarasov, Denis and Kolesnikov, Sergey},
  booktitle={International conference on machine learning},
  pages={26228--26244},
  year={2023},
  organization={PMLR}
}

@inproceedings{optidice_lee2021,
  title={Optidice: Offline policy optimization via stationary distribution correction estimation},
  author={Lee, Jongmin and Jeon, Wonseok and Lee, Byungjun and Pineau, Joelle and Kim, Kee-Eung},
  booktitle={International Conference on Machine Learning},
  pages={6120--6130},
  year={2021},
  organization={PMLR}
}

@article{dualrl_sikchi2024,
  title={Dual rl: Unification and new methods for reinforcement and imitation learning},
  author={Sikchi, Harshit and Zheng, Qinqing and Zhang, Amy and Niekum, Scott},
  journal={arXiv preprint arXiv:2302.08560},
  year={2023}
}

@article{dt_chen2021,
  title={Decision transformer: Reinforcement learning via sequence modeling},
  author={Chen, Lili and Lu, Kevin and Rajeswaran, Aravind and Lee, Kimin and Grover, Aditya and Laskin, Misha and Abbeel, Pieter and Srinivas, Aravind and Mordatch, Igor},
  journal={Advances in neural information processing systems},
  volume={34},
  pages={15084--15097},
  year={2021}
}

@article{tt_janner2021,
  title={Offline reinforcement learning as one big sequence modeling problem},
  author={Janner, Michael and Li, Qiyang and Levine, Sergey},
  journal={Advances in neural information processing systems},
  volume={34},
  pages={1273--1286},
  year={2021}
}

@article{diffuser_janner2022,
  title={Planning with diffusion for flexible behavior synthesis},
  author={Janner, Michael and Du, Yilun and Tenenbaum, Joshua B and Levine, Sergey},
  journal={arXiv preprint arXiv:2205.09991},
  year={2022}
}

@article{flowql_park2025,
  title={Flow q-learning},
  author={Park, Seohong and Li, Qiyang and Levine, Sergey},
  journal={arXiv preprint arXiv:2502.02538},
  year={2025}
}

@article{synthER_lu2023,
  title={Synthetic experience replay},
  author={Lu, Cong and Ball, Philip and Teh, Yee Whye and Parker-Holder, Jack},
  journal={Advances in Neural Information Processing Systems},
  volume={36},
  pages={46323--46344},
  year={2023}
}

@article{gta_lee2024,
  title={Gta: Generative trajectory augmentation with guidance for offline reinforcement learning},
  author={Lee, Jaewoo and Yun, Sujin and Yun, Taeyoung and Park, Jinkyoo},
  journal={Advances in Neural Information Processing Systems},
  volume={37},
  pages={56766--56801},
  year={2024}
}

@article{diffstitch_li2024,
  title={Diffstitch: Boosting offline reinforcement learning with diffusion-based trajectory stitching},
  author={Li, Guanghe and Shan, Yixiang and Zhu, Zhengbang and Long, Ting and Zhang, Weinan},
  journal={arXiv preprint arXiv:2402.02439},
  year={2024}
}

@inproceedings{rtdiff_yang2025,
  title={Rtdiff: Reverse trajectory synthesis via diffusion for offline reinforcement learning},
  author={Yang, Q and Wang, YX},
  booktitle={International Conference on Learning Representations (ICLR)},
  year={2025}
}

@article{bitrajdiff_qing2025,
  title={BiTrajDiff: Bidirectional Trajectory Generation with Diffusion Models for Offline Reinforcement Learning},
  author={Qing, Yunpeng and Chen, Shuo and Chi, Yixiao and Liu, Shunyu and Lin, Sixu and Zou, Changqing},
  journal={arXiv preprint arXiv:2506.05762},
  year={2025}
}

@article{mbts_hepburn2022,
  title={Model-based trajectory stitching for improved offline reinforcement learning},
  author={Hepburn, Charles A and Montana, Giovanni},
  journal={arXiv preprint arXiv:2211.11603},
  year={2022}
}

@article{scots_lee2025,
  title={State-Covering Trajectory Stitching for Diffusion Planners},
  author={Lee, Kyowoon and Choi, Jaesik},
  journal={arXiv preprint arXiv:2506.00895},
  year={2025}
}

@article{pgd_jackson2024,
  title={Policy-guided diffusion},
  author={Jackson, Matthew Thomas and Matthews, Michael Tryfan and Lu, Cong and Ellis, Benjamin and Whiteson, Shimon and Foerster, Jakob},
  journal={arXiv preprint arXiv:2404.06356},
  year={2024}
}

@article{MOPO_2020,
  title={Mopo: Model-based offline policy optimization},
  author={Yu, Tianhe and Thomas, Garrett and Yu, Lantao and Ermon, Stefano and Zou, James Y and Levine, Sergey and Finn, Chelsea and Ma, Tengyu},
  journal={Advances in Neural Information Processing Systems},
  volume={33},
  pages={14129--14142},
  year={2020}
}

@article{morel_kidambi2020,
  title={Morel: Model-based offline reinforcement learning},
  author={Kidambi, Rahul and Rajeswaran, Aravind and Netrapalli, Praneeth and Joachims, Thorsten},
  journal={Advances in neural information processing systems},
  volume={33},
  pages={21810--21823},
  year={2020}
}

@article{combo_yu2021,
  title={Combo: Conservative offline model-based policy optimization},
  author={Yu, Tianhe and Kumar, Aviral and Rafailov, Rafael and Rajeswaran, Aravind and Levine, Sergey and Finn, Chelsea},
  journal={Advances in neural information processing systems},
  volume={34},
  pages={28954--28967},
  year={2021}
}

@article{bats_char2022,
  title={Bats: Best action trajectory stitching},
  author={Char, Ian and Mehta, Viraj and Villaflor, Adam and Dolan, John M and Schneider, Jeff},
  journal={arXiv preprint arXiv:2204.12026},
  year={2022}
}

@article{leq_park2024,
  title={Model-based Offline Reinforcement Learning with Lower Expectile Q-Learning},
  author={Park, Kwanyoung and Lee, Youngwoon},
  journal={arXiv preprint arXiv:2407.00699},
  year={2024}
}

@article{ho2020ddpm,
  title={Denoising diffusion probabilistic models},
  author={Ho, Jonathan and Jain, Ajay and Abbeel, Pieter},
  journal={Advances in neural information processing systems},
  volume={33},
  pages={6840--6851},
  year={2020}
}

@article{song2021sde,
  title={Score-based generative modeling through stochastic differential equations},
  author={Song, Yang and Sohl-Dickstein, Jascha and Kingma, Diederik P and Kumar, Abhishek and Ermon, Stefano and Poole, Ben},
  journal={arXiv preprint arXiv:2011.13456},
  year={2020}
}

@inproceedings{ssd_kim2024,
  title={Stitching sub-trajectories with conditional diffusion model for goal-conditioned offline rl},
  author={Kim, Sungyoon and Choi, Yunseon and Matsunaga, Daiki E and Kim, Kee-Eung},
  booktitle={Proceedings of the AAAI Conference on Artificial Intelligence},
  volume={38},
  number={12},
  pages={13160--13167},
  year={2024}
}

@article{compdiffuser_luo2025,
  title={Generative trajectory stitching through diffusion composition},
  author={Luo, Yunhao and Mishra, Utkarsh A and Du, Yilun and Xu, Danfei},
  journal={arXiv preprint arXiv:2503.05153},
  year={2025}
}

@article{mbrcsl_zhou2023,
  title={Free from bellman completeness: Trajectory stitching via model-based return-conditioned supervised learning},
  author={Zhou, Zhaoyi and Zhu, Chuning and Zhou, Runlong and Cui, Qiwen and Gupta, Abhishek and Du, Simon Shaolei},
  journal={arXiv preprint arXiv:2310.19308},
  year={2023}
}

@inproceedings{gas_baek2025,
title={Graph-Assisted Stitching for Offline Hierarchical Reinforcement Learning},
author={Seungho Baek and taegeon park and Jongchan Park and Seungjun Oh and Yusung Kim},
booktitle={Forty-second International Conference on Machine Learning},
year={2025},
url={https://openreview.net/forum?id=73EwiOrN8W}
}

@article{levine2020offline,
  title={Offline reinforcement learning: Tutorial, review, and perspectives on open problems},
  author={Levine, Sergey and Kumar, Aviral and Tucker, George and Fu, Justin},
  journal={arXiv preprint arXiv:2005.01643},
  year={2020}
}

@inproceedings{agarwal2020optimistic,
  title={An optimistic perspective on offline reinforcement learning},
  author={Agarwal, Rishabh and Schuurmans, Dale and Norouzi, Mohammad},
  booktitle={International conference on machine learning},
  pages={104--114},
  year={2020},
  organization={PMLR}
}

@article{park2024ogbench,
  title={Ogbench: Benchmarking offline goal-conditioned rl},
  author={Park, Seohong and Frans, Kevin and Eysenbach, Benjamin and Levine, Sergey},
  journal={arXiv preprint arXiv:2410.20092},
  year={2024}
}

@article{fu2020d4rl,
  title={D4RL: Datasets for Deep Data-Driven Reinforcement Learning},
  author={Fu, Justin and Kumar, Aviral and Nachum, Ofir and Tucker, George and Levine, Sergey},
  journal={arXiv preprint arXiv:2004.07219},
  year={2020}
}

@article{hilbert_paper,
  title={Foundation policies with hilbert representations},
  author={Park, Seohong and Kreiman, Tobias and Levine, Sergey},
  journal={arXiv preprint arXiv:2402.15567},
  year={2024}
}

@article{tldr_paper,
  title={Tldr: Unsupervised goal-conditioned rl via temporal distance-aware representations},
  author={Bae, Junik and Park, Kwanyoung and Lee, Youngwoon},
  journal={arXiv preprint arXiv:2407.08464},
  year={2024}
}

@inproceedings{cudc_paper,
  title={Cudc: A curiosity-driven unsupervised data collection method with adaptive temporal distances for offline reinforcement learning},
  author={Sun, Chenyu and Qian, Hangwei and Miao, Chunyan},
  booktitle={Proceedings of the AAAI Conference on Artificial Intelligence},
  volume={38},
  number={13},
  pages={15145--15153},
  year={2024}
}

@article{tempdata_paper,
  title={Temporal Distance-aware Transition Augmentation for Offline Model-based Reinforcement Learning},
  author={Lee, Dongsu and Kwon, Minhae},
  journal={arXiv preprint arXiv:2505.13144},
  year={2025}
}

@inproceedings{li2024towards_neurips,
  title     = {Towards an Information Theoretic Framework of Context-Based Offline Meta-Reinforcement Learning},
  author    = {Lanqing Li and Hai Zhang and Xinyu Zhang and Shatong Zhu and Yang Yu and Junqiao Zhao and Pheng-Ann Heng},
  booktitle = {Advances in Neural Information Processing Systems},
  volume    = {37},
  year      = {2024},
  url       = {https://proceedings.neurips.cc/paper_files/paper/2024/file/8a30aba6514b56d02976f49797f6338a-Paper-Conference.pdf}
}
% \input{sections/6_theory}
% \newpage
% \newpage

% \input{sections/6_theory_final}

\clearpage

\section*{Appendix~A: Detailed Algorithms}

\begin{algorithm}[ht!]
    \caption{Dynamics-Guided Stitch Planner Training}
    \label{alg:planner_training}
    \begin{algorithmic}[1]
    \State Assume pretrained dynamics model $\theta_{\text{dyn}}$ is fixed
    \State Initialize stitch planner $\theta_{\pi}$
    \Repeat
        \State Sample clean trajectory $(\tau^s, \tau^a)$ from dataset
        \State Sample diffusion timestep $t \sim \mathcal{U}(\{1, \dots, T\})$
        \State Corrupt actions to obtain noisy $\tau_{t}^a$ via forward diffusion
        \State {Compute rollout deviation using eq.11}
        \State {Compute self-conditioned loss $\mathcal{L}_{\text{self-cond}}$ using  eq.15}
        \State {Compute regularization loss $\mathcal{L}_{\text{reg}}$ using eq.14}
        \State Compute total loss $\mathcal{L}(\theta_{\pi}) = \mathcal{L}_{\text{self-cond}} + \mathcal{L}_{\text{reg}}$; update $\theta_{\pi}$
    \Until{converged}
    \State \textbf{Return} trained stitch planner $\theta_{\pi}$
    \end{algorithmic}
    \end{algorithm}
    
    \begin{algorithm}[ht!]
    \caption{Dynamics-Guided Stitch Planner Inference}
    % \label{alg:planner_inference}
    \begin{algorithmic}[1]
    \State Given masked stitching sequence $\tau^{s}_{m}$
    \State Initialize noisy action sequence $\hat{\tau}_{T}^{a} \sim \mathcal{N}(0,I)$
    \State rollout deviation $\delta(\tilde{\tau}_{0}^{a,(T+1)})=0$
    \For{$t = T$ down to $1$}
        \State Compute rollout deviation $\delta(\tilde{\tau}_{0}^{a,(t+1)})$ using eq.11
        \State {Sample $\hat{\tau}_{t-1}^{a}$ using eq.9}
    \EndFor
    \State \textbf{Return} refined action sequence $\hat{\tau}_{0}^{a}$
    \end{algorithmic}
    \end{algorithm}
    
    \begin{algorithm}[ht!]
    \caption{Policy Training with Adaptive Weighted Trajectory Stitch Augmentation}
    \label{alg:policy_training}
    \begin{algorithmic}[1]
    \State Given pretrained dynamics model $\theta_{\text{dyn}}$, stitch planner $\theta_{\pi}$, and offline dataset $\mathcal{D}$
    \State Initialize policy $\pi_{\theta}$, critic $Q_{\phi}$, and stitched buffer $\mathcal{B}_{\text{stitch}}=\emptyset$
    \For{each offline RL training iteration}
        \State Perform temporal-space stitch target selection on $\mathcal{D}$ to generate source-target state pairs
        \State Use stitch planner $\theta_{\pi}$ to generate stitched actions $\hat{\tau}_{0}^{a}$
        \State Use dynamics model $\theta_{\text{dyn}}$ to roll out stitched states $\hat{\tau}_{0}^{s}$ given $s_0$ and $\hat{\tau}_{0}^{a}$
        \State Form stitched trajectory $\hat{\tau}_i = (\hat{\tau}_{0}^{s}, \hat{\tau}_{0}^{a})$
        \State Sample mini-batch from $\mathcal{D}$ and buffer $\mathcal{B}_{\text{stitch}}$
        \State Update policy $\pi_{\theta}$ and critic $Q_{\phi}$ using offline RL algorithm
    \EndFor
    \State \textbf{Return} optimized policy $\pi_{\theta}$
    \end{algorithmic}
    \end{algorithm}
\begin{algorithm}[ht!]
    
\caption{TDR-based Sequence Filter}
\label{alg:mask_filter}
\begin{algorithmic}[1]
\Require Stitching sequence $\tau^{s}_{\text{m}}$; TDR encoder $\psi$; sample count $k$; threshold $\Delta_{\text{thresh}}$
\Ensure \textbf{keep} $\in\{\text{true},\text{false}\}$
\State $\mathcal{B} \gets \emptyset$       \Comment{store distance biases}
\For{$t = 1$ \textbf{to} $k$}
    \State Randomly sample state pairs $(s_m, s_n)$ from $\tau^{s}_{\text{m}}$
    \State $d_{\text{step}} \gets |m-n|$  \Comment{temporal distance}
    \State $d_{\text{TDR}} \gets \|\psi(s_m)-\psi(s_n)\|_2$
    \State $\mathcal{B} \gets \mathcal{B} \cup \{|d_{\text{step}}-d_{\text{TDR}}|\}$
\EndFor
\State $\mathbb{E}[\Delta_d] \gets \frac{1}{k}\sum_{b \in \mathcal{B}} b$
\Return $\textbf{keep} \gets (\mathbb{E}[\Delta_d] \le \Delta_{\text{thresh}})$
\end{algorithmic}
\end{algorithm}

%\vspace{3cm}

\section*{Appendix~B: Environments and Model Implementation}

\subsection*{B.1\quad Environments and Datasets}

We use the OGBench and D4RL datasets for evaluation. For OGBench maze, we use stitch dataset to validate our method. Figure shows the difference between stitch dataset and navigate and explore dataset, which consists of short trajectories that RL must stitch between trajectories to enable effective planning, which causing the low performance of the offline RL baselines without augmentation. The performance improvement in all environments shows our augmentation is effective.

\subsubsection*{Environments}
\begin{figure*}[t!]
    \centering
    \includegraphics[width=1.0\textwidth]{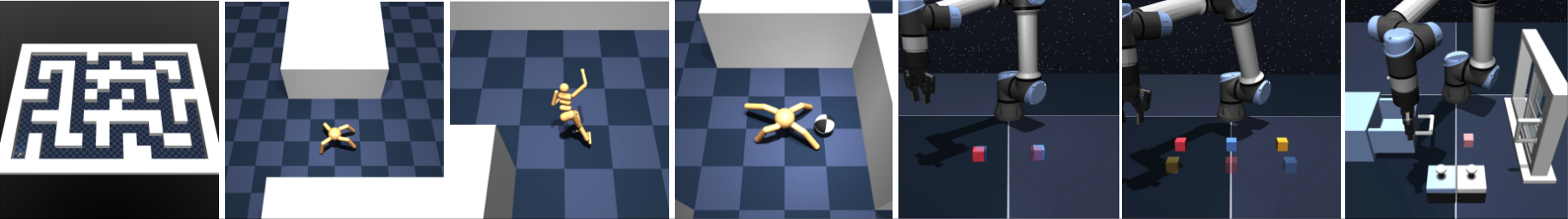}
    \caption{OGBench environments}
    \vspace{-0.5cm}
    \label{fig:stitch_dataset}
\end{figure*}
\paragraph{Antmaze.}
In this dataset, an agent controls a quadruped Ant robot with 8 degrees of freedom (DoF). The objective is to navigate through various maze configurations to reach a designated goal location, requiring simultaneous mastery of high-level pathfinding and low-level locomotion control. The environments are provided in three distinct maze sizes: medium, large, and giant, with larger mazes specifically crafted to evaluate extensive long-horizon planning capabilities. Observations are represented by a 29-dimensional state vector, capturing the robot's two-dimensional position (x-y coordinates) and detailed joint-related information. Simulations are executed within the MuJoCo physics simulator.
\vspace{-0.2cm}
\paragraph{Humanoidmaze.}
This dataset introduces increased complexity by utilizing a Humanoid robot with 17 degrees of freedom (DoF). The task involves navigating medium and large mazes, significantly testing long-horizon planning and sophisticated locomotion control due to the robot’s more intricate dynamics. The state observation consists of a 376-dimensional vector, capturing comprehensive joint positions, velocities, and various proprioceptive signals. All simulations are performed using the MuJoCo physics engine.
\vspace{-0.2cm}

\paragraph{Antsoccer-arena.}
This environment integrates manipulation and locomotion, tasking an Ant robot with 8 DoF to simultaneously dribble a ball and navigate within an arena setting. This task evaluates the agent’s capacity for coupled object manipulation and precise locomotion. Observations are state-based and include a 40-dimensional vector covering the agent’s positional information, joint states, and interactions with the ball. The environment is simulated in MuJoCo.
\vspace{-0.2cm}

\paragraph{Scene-play.}
The Scene-play dataset examines sequential decision-making skills requiring complex reasoning about multiple interactive objects, including a drawer, window, button locks, and a cube. Data is collected through open-loop, non-Markovian scripted interactions, mimicking play-like behavior with diverse everyday objects. Observations are state-based, capturing comprehensive object poses and robotic joint states. Simulated in MuJoCo.

\paragraph{Cube-single-play.}
This dataset assesses an agent’s capability to manipulate a single cube using flexible pick-and-place strategies. Data collection involves randomized open-loop scripted interactions, emphasizing robust generalization across various manipulation tasks. Observations include detailed positions and orientations of the cube alongside robotic joint states. The simulations are conducted using MuJoCo.

\paragraph{Cube-double-play.}
Building upon single-object manipulation, this dataset introduces complexity by requiring the coordinated handling and stacking of two cubes. Data is collected via open-loop scripted interactions, emphasizing sequential reasoning and compositional task generalization. Observations consist of positions, orientations of both cubes, and detailed robotic joint states. The environment simulations utilize MuJoCo.

\subsubsection*{Dataset Types}

\paragraph{Stitching.}
The stitching dataset type specifically evaluates the agent’s proficiency in assembling partial trajectories into cohesive, optimal plans. Data comprises short goal-reaching trajectories generated by noisy expert policies, navigating to randomly sampled intermediate goals.
\begin{figure}[h!]
    \centering
    \includegraphics[width=0.5\textwidth]{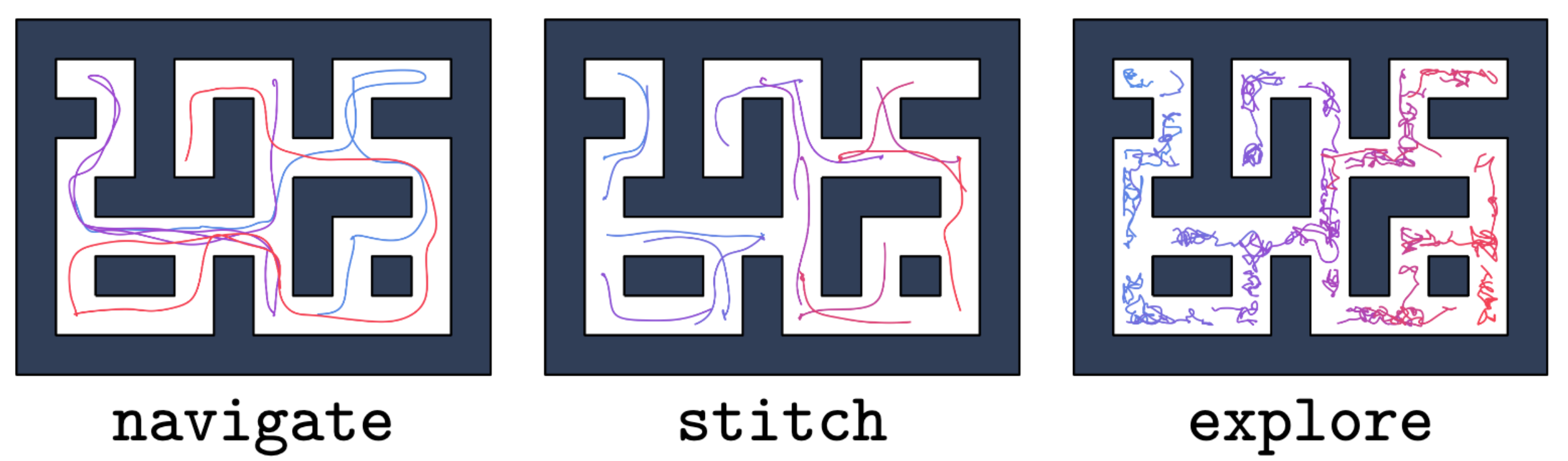}
    \caption{Difference between stitch dataset and navigate and explore dataset.}
\end{figure}
\paragraph{Manipulation.}
The manipulation dataset type evaluates sequential decision-making capabilities, emphasizing reasoning about past actions to inform future plans. The play-style dataset employs open-loop, non-Markovian scripted interactions to simulate diverse object manipulations. The partial dataset derives from human demonstrations originally captured using a VR teleoperation interface.

\subsubsection*{Goal Specification for Evaluation}
Goal specification and evaluation follow the established OGBench protocol, utilizing default tasks and assessing agent performance across 8 random seeds per environment.

\subsubsection*{Evaluation Metric}
Performance evaluation utilizes the standard success rate metric.

\subsection*{B.2\quad Model Implementation}

\paragraph*{B.2.1\quad Base RL Implementation}

We implement two state-of-the-art offline RL algorithms as base learners for ASTRO evaluation:

{IQL (Implicit Q-Learning)}: A conservative one-step algorithm that performs advantage-weighted behavior cloning. IQL learns a value function $V(s)$ and advantage function $A(s,a)$ through implicit Q-learning, then extracts the policy via:
\begin{equation}
\pi(a|s) \propto \exp(\beta A(s,a))
\end{equation}
where $\beta$ is a temperature parameter controlling conservatism. IQL avoids distributional shift by staying close to the behavior policy while maximizing expected returns.

{FQL (Flow Q-Learning)}: A more expressive algorithm that uses flow-matching-based action sampling. FQL learns a flow model that directly parameterizes the policy distribution, enabling more flexible action generation compared to IQL's conservative approach. The flow model is trained to match the optimal action distribution through continuous normalizing flows, providing better exploration of the action space while maintaining policy improvement guarantees.

Both algorithms are implemented with identical network architectures (3-layer MLPs with 256 hidden units) and training protocols (learning rate $3\times10^{-4}$, batch size 256, 1M gradient steps) to ensure fair comparison across augmentation methods.

We follow the hyperparameter settings from FQL for all environments and algorithms, shown in table~\ref{tab:iql-fql-hparams}.

\begin{table*}[t]
    \centering
    \caption{IQL and FQL hyperparameters used in our experiments.}
    \label{tab:iql-fql-hparams}
    \begin{tabular}{lccc}
    \toprule
    \textbf{Task} & \textbf{IQL $\boldsymbol{\alpha}$} & \textbf{FQL $\boldsymbol{\alpha}$} & \textbf{Discount $\boldsymbol{\gamma}$} \\
    \midrule
    ant-large-v0 & 10   & 10   & 0.99 \\
    ant-giant-v0 & 10   & 10   & 0.995 \\
    humanoid-medium-v0 & 10 & 30 & 0.995 \\
    humanoid-large-v0  & 10 & 30 & 0.995 \\
    antsoccer-arena-v0 & 1  & 10 & 0.995 \\
    scene-v0           & 10 & 100 & 0.99 \\
    cube-single-v0     & 1  & 300 & 0.99 \\
    cube-double-v0     & 0.3 & 100 & 0.99 \\
    \textbf{D4RL avg (6 ant-mazes)} & -- & 10 (umaze/medium), 3 (large) & 0.99 \\
    \bottomrule
    \end{tabular}
\end{table*}

\begin{table*}[ht]
    \centering
    \caption{Horizon lengths used for each environment.}
    \label{tab:horizon-settings}
    \begin{tabular}{llc}
    \toprule
    \textbf{Benchmark} & \textbf{Environment} & \textbf{Horizon} \\
    \midrule
    OGBench AntMaze Stitch   
                             & antmaze-large-v0          & 180 \\
                             & antmaze-giant-v0          & 180 \\
    OGBench HumanoidMaze Stitch & humanoid-medium-v0     & 340 \\
                                & humanoid-large-v0      & 340 \\
    OGBench AntSoccer Stitch & antsoccer-arena-v0        & 180 \\

    OGBench AntSoccer Stitch & antsoccer-arena-v0        & 180 \\
    OGBench scene & -        & 180 \\
    OGBench cube & single        & 180 \\
                 & double        & 180 \\
    D4RL AntMaze Stitch      & - & 180 \\
    \bottomrule
\end{tabular}
\vspace{-5mm}
\end{table*}

All experiments use identical random seeds and evaluation protocols across methods for fair comparison.

\paragraph*{B.2.2\quad Diffusion Implementation}

\paragraph{Network Architecture.}  
For both stitch planner and dynamics model, we use a diffusion transformer architecture based on DiT for sequence modeling in high-dimensional continuous control. Specifically, we adopt the publicly released DiT model architecture (from facebookresearch) as the denoising network within our diffusion framework. The network leverages a transformer backbone designed for conditional generation, allowing it to scale to long horizons and complex dynamics. To allow different field of state and action, we use unified concat of state and action sequence to form as trajectory sequence.

\vspace{-0.2cm}

\paragraph{Horizon.}  
We adopt task-specific fixed horizons to align with the complexity and planning requirements of each environment. The specific horizon lengths used during training and inference are summarized in Table~\ref{tab:horizon-settings}.

\vspace{0.5em}
\begin{table*}[t]
\centering
\caption{Ablation results on \texttt{antmaze-large-v0}. Policy performance and selection latency (in seconds) are reported.}
\label{tab:ablation-all}
\begin{subtable}[t]{0.34\textwidth}
\centering\small
\caption{Filter Threshold}
\begin{tabular}{lcc}
\toprule
\textbf{Threshold} & \textbf{Performance} & \textbf{Time} \\
\midrule
5         & 48.9 & 2.1 \\
\textbf{3}         & \textbf{57.3} & \textbf{4.5} \\
2         & 58.2 & 12.4 \\
\bottomrule
\end{tabular}
\end{subtable}
\hfill
\begin{subtable}[t]{0.34\textwidth}
\centering\small
\caption{Chain Length}
\begin{tabular}{lcc}
\toprule
\textbf{Length} & \textbf{Success (\%)} & \textbf{Time} \\
\midrule
3 & 52.1 & 2.4 \\
\textbf{5} & \textbf{57.3} & \textbf{6.5} \\
7 & 59.6 & 25.6 \\
\bottomrule
\end{tabular}
\end{subtable}
\hfill
\begin{subtable}[t]{0.15\textwidth}
\centering\small
\caption{Horizon $L$}
\begin{tabular}{lc}
\toprule
\textbf{$L$} & \textbf{MSE $\downarrow$} \\
\midrule
120 & 0.673 \\
\textbf{160} & \textbf{0.657} \\
200 & 0.698 \\
\bottomrule
\end{tabular}
\end{subtable}
\hfill
\begin{subtable}[t]{0.15\textwidth}
\centering\small
\caption{Mask $M/l$}
\begin{tabular}{lc}
\toprule
\textbf{$M/l$} & \textbf{MSE $\downarrow$} \\
\midrule
1/3 & 0.672 \\
\textbf{1/1} & \textbf{0.657} \\
3/1 & 0.715 \\
\bottomrule
\end{tabular}
\end{subtable}
\vspace{-0.5cm}
\end{table*}

\paragraph*{B.2.3\quad TDR Selection Implementation}

We adopt a goal relabeling strategy similar to prior works, with a key modification: we explicitly exclude the trivial case where the current state and goal are identical ($s = g$), since our temporal distance formulation satisfies $V(s, s) = 0$ by construction. 

At each training step, the goal state $g$ is sampled using the following mixture strategy. With probability $0.625$, $g$ is sampled from a future state along the same trajectory, where the offset follows a geometric distribution over time indices. With probability $0.375$, $g$ is uniformly sampled from the full replay buffer, enabling global goal matching. This sampling scheme ensures that $s = g$ is never selected, and reflects the original hyperparameter design that balances local temporal reasoning with global generalization.

The temporal distance predictor $V(s, g)$ is modeled as a feed-forward multilayer perceptron (MLP) with architecture $(512, 512, 512)$, outputting a 32-dimensional latent embedding or 64-dimensional latent embedding(for humanoidmaze). To stabilize training and improve representation consistency, we apply Layer Normalization to the MLP hidden activations.

For selecting the masked stitching sequence, we choose a unified stitching chain of 5 to different tasks, and use \(l/M\) = 1 to facilatate balance mask and sub-trajectory information. Ablation against hyperparameters is shown in Appendix D.
\subsubsection*{B.2.4\quad Baselines Implementation}

For fair comparison, we use the same implementation of our DiT to evaluate Diffstitch and SynthER. We tune the low-to-high reward strategy to noised sampling to ensure balanced augmentation for DiffStitch in sparse reward environments. For other hyperparameters such as Euclidean threshold, we follow the original implementation.

\section*{Appendix~C: Metrics for Geometric Consistency}

To quantify the geometric quality of stitched trajectories, we employ three metrics that capture local smoothness, directional coherence and dynamics consistency:

\textbf{Directional Change ($\Delta \theta$).}  
We compute the change in heading angle between consecutive segments to measure abrupt directional shifts. Given three consecutive positions $s_{t-1}, s_t, s_{t+1}$, the angle difference is computed as:
\begin{equation}
\Delta \theta_t = \angle(s_t - s_{t-1}, s_{t+1} - s_t),
\end{equation}
where $\angle(\cdot, \cdot)$ computes the signed angle between two 2D vectors.

\textbf{Trajectory Curvature.}  
We further measure the local curvature to assess how sharply the agent turns along the path. The curvature $\kappa_t$ at timestep $t$ is defined by:
\begin{equation}
\kappa_t = \frac{\|s_{t+1} - 2s_t + s_{t-1}\|}{\|s_{t+1} - s_{t}\|^2 + \|s_t - s_{t-1}\|^2},
\end{equation}

where $\|\cdot\|$ denotes the Euclidean norm. This metric penalizes high-frequency oscillations and is sensitive to trajectory smoothness.

\textbf{Dynamics Violation Rate.}  
To evaluate the physical feasibility of stitched transitions, we introduce the \textit{dynamics violation rate}, which quantifies how often a transition $(s, a, s')$ violates the underlying environment dynamics.

Given a stitched pair $(s, s')$ (without known action), we sample $n$ random actions $\{a_i\}_{i=1}^n$ from the environment's action space and evaluate whether any of them can plausibly generate $s'$ from $s$. Specifically, we define the violation criterion as:

\begin{equation}
\mathcal{V}(s, s') =
\begin{cases}
1, & \text{if } \forall a_i \in \mathcal{A},\ \|f(s, a_i) - s'\| > \delta, \\
0, & \text{otherwise},
\end{cases}
\end{equation}

where $f(s, a_i)$ is the next state predicted by stepping the environment forward with action $a_i$, and $\delta$ is a small threshold for matching the next state.

The dynamics violation rate is computed as the average across all stitched transitions:
\begin{equation}
\text{Violation Rate} = \frac{1}{N} \sum_{j=1}^N \mathcal{V}(s_j, s'_j),
\end{equation}

where $N$ is the number of stitched transitions being evaluated.

\section*{Appendix~D: Ablation Study}

In this section, we conduct comprehensive ablation studies on critical design choices that directly impact ASTRO's ability to generate distributionally novel and dynamics-consistent trajectories. All experiments are conducted on \texttt{antmaze-large-v0}, a challenging long-horizon navigation task with fragmented demonstration data that exemplifies the core problems addressed by ASTRO and FQL as RL base agent.

\subsection*{D.1\quad Ablation on TDR-based Stitch Target Selection}

We first analyze how TDR design choices affect the identification of distinct and reachable stitch targets, which is fundamental to ASTRO's ability to create novel trajectory connections. The time below is the average selection time of 10 samples.

\textbf{Filter Threshold.} The TDR filter threshold determines the quality of selected stitch targets by controlling how strict the temporal-distance constraints are. Tighter thresholds (lower values) ensure that only highly compatible state pairs are considered for stitching, leading to more dynamics-consistent augmented trajectories and improved policy performance. However, this comes at the cost of significantly increased computational overhead due to more stringent filtering requirements. A threshold of 3 provides the optimal trade-off between trajectory quality and computational efficiency, as shown in Table~\ref{tab:ablation-all}(a).

\textbf{Chain Length.} Longer subtrajectory Chain Length enable more sophisticated trajectory compositions and can potentially discover more diverse stitching paths through the state space. While longer Chain Length provide marginal performance improvements by enabling more complex trajectory augmentations, they suffer from exponential increases in selection latency due to the combinatorial growth of possible chain configurations. A Chain Length of 5 achieves the best balance between augmentation diversity and practical computational constraints, as demonstrated in Table~\ref{tab:ablation-all}(b).

\subsection*{D.2\quad Ablation on Dynamics-Guided Diffusion Planning}

Next, we analyze hyperparameters that affect the quality of dynamics-guided action sequence generation, which is crucial for ensuring the feasibility and reachability of stitched trajectories. We use generated rollout state MSE as metric.

\textbf{Diffusion Horizon $L$.} The trajectory horizon length directly impacts the model's ability to capture long-range dependencies and generate coherent action sequences for trajectory stitching. Moderate horizons ($L=160$) provide sufficient context for effective dynamics modeling without overfitting to specific trajectory patterns. Excessively long horizons can lead to degraded performance due to increased model complexity and potential overfitting to suboptimal trajectory segments, as evidenced in Table~\ref{tab:ablation-all}(c).

\textbf{Subtrajectory Masking Ratio $M/l$.} This ratio controls the balance between global trajectory structure preservation and local action sequence generation during diffusion training. A balanced setting ($M/l=1$) ensures that the model learns to generate action sequences that are both locally coherent and globally consistent with the overall trajectory dynamics. Extreme ratios either under-constrain the generation process (leading to dynamics violations) or over-constrain it (strict target reaching goal), as shown in Table~\ref{tab:ablation-all}(d).

\section*{Appendix~E: Additional visualizations}

Here we present some selection examples for visualization, showing TDR-based robust and distinct selection for trajectory stitching.

\begin{figure*}[h!]
    \centering
    \includegraphics[width=0.9\textwidth]{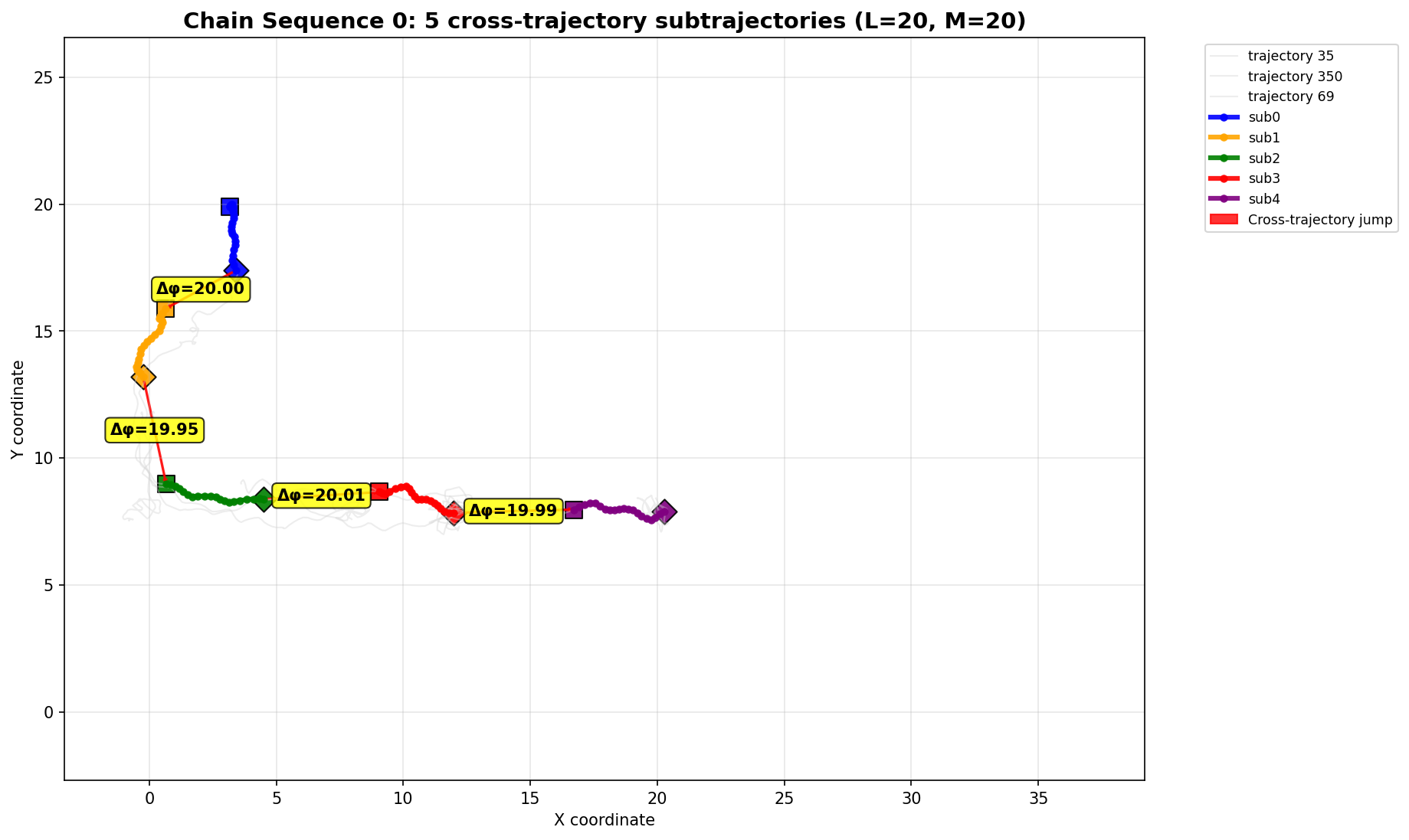}
    \caption{Visualization of selection results 0.}
\end{figure*}

\begin{figure*}[h!]
    \centering
    \includegraphics[width=0.9\textwidth]{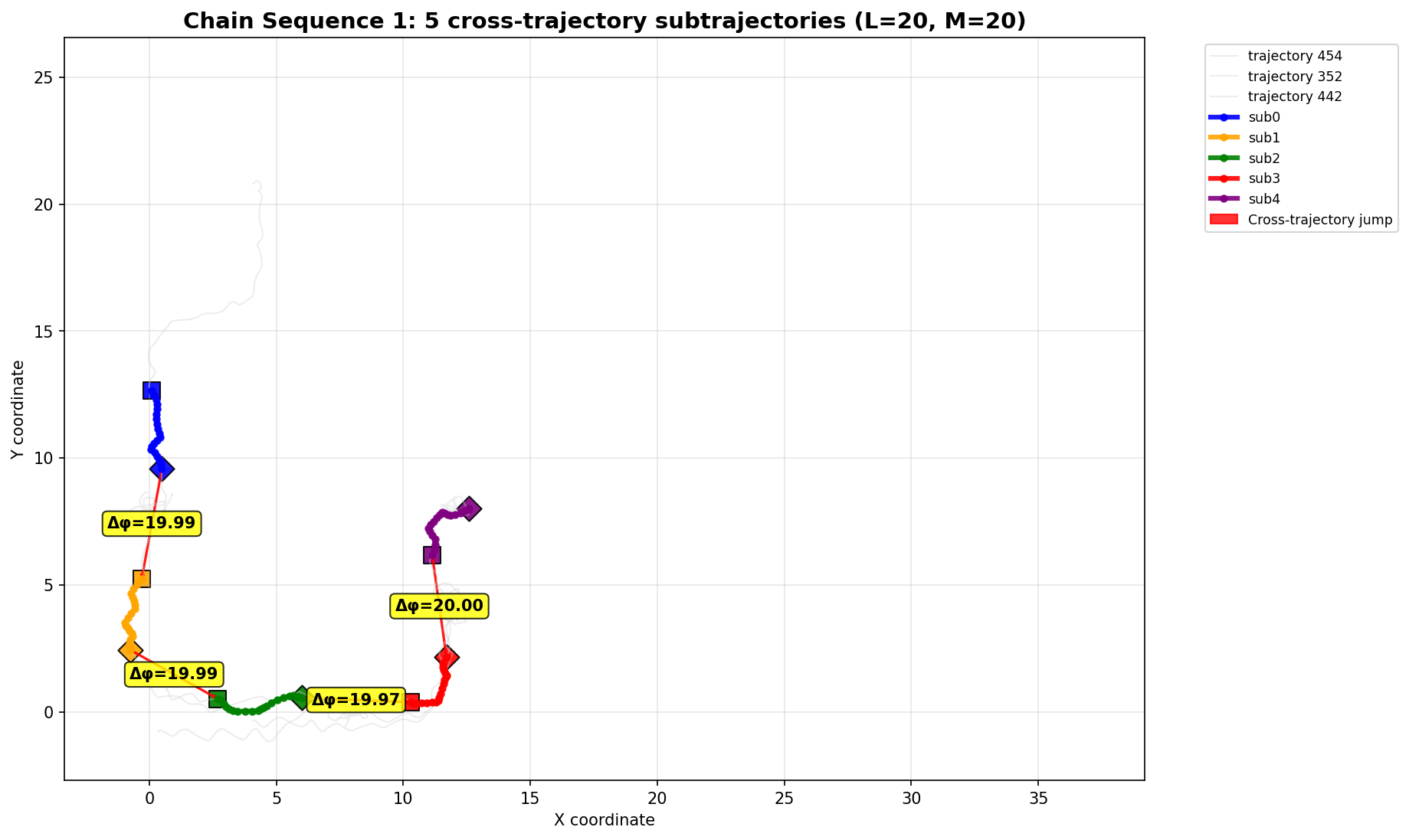}
    \caption{Visualization of selection results 1.}

\end{figure*}

\begin{figure*}[h!]
    \centering
    \includegraphics[width=0.9\textwidth]{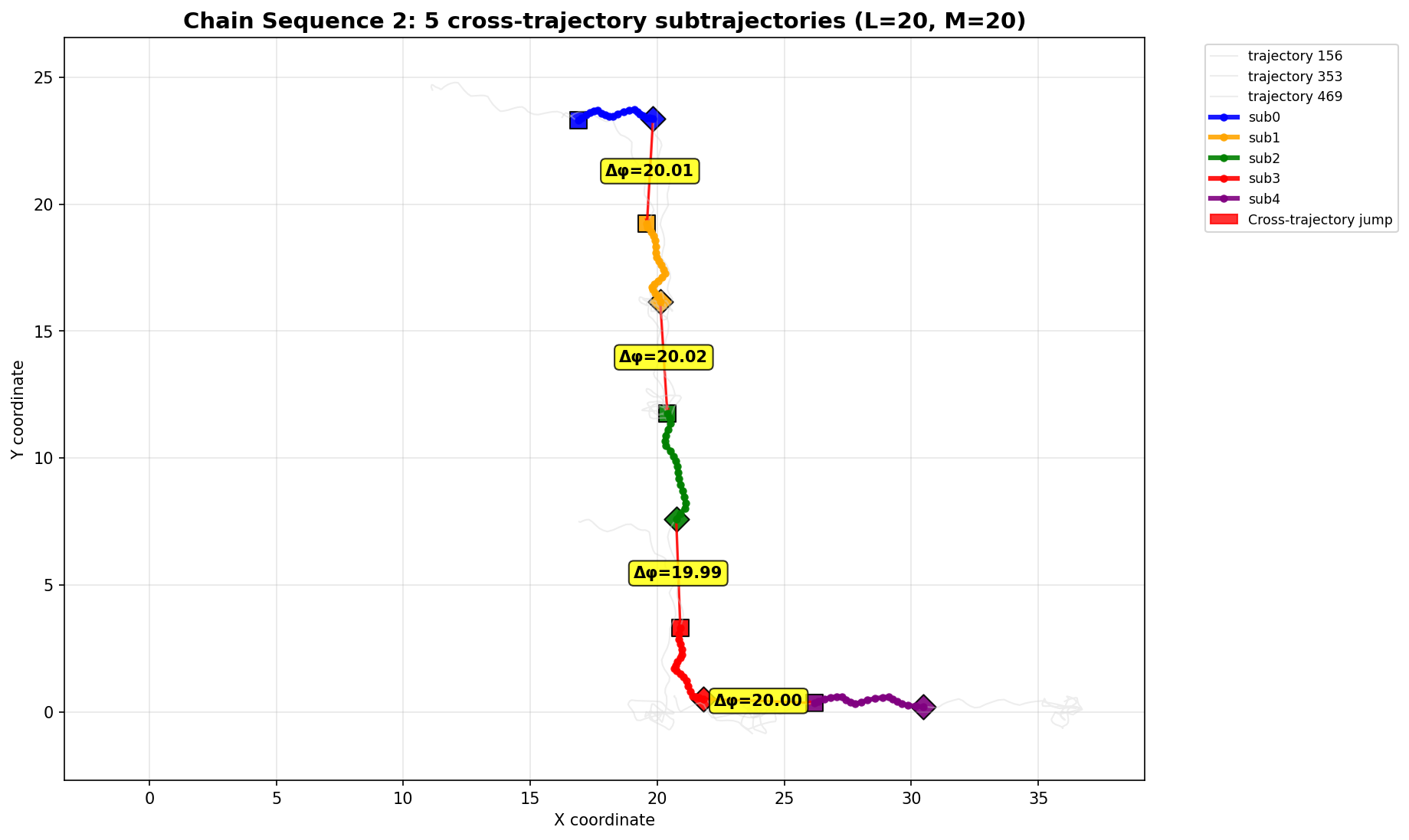}
    \caption{Visualization of selection results 2.}

\end{figure*}

\begin{figure*}[h!]
    \centering
    \includegraphics[width=0.9\textwidth]{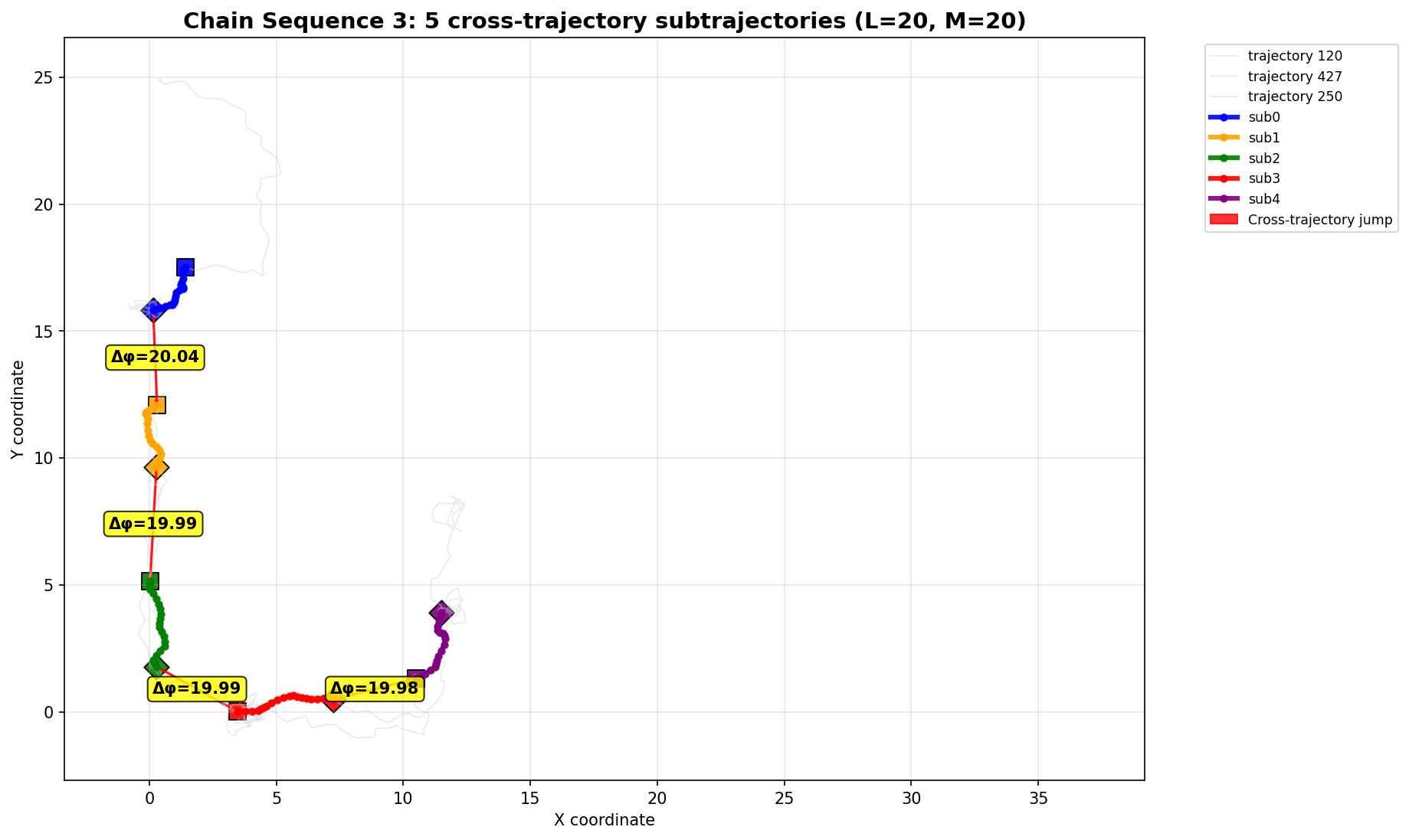}
    \caption{Visualization of selection results 3.}

\end{figure*}

\begin{figure*}[h!]
    \centering
    \includegraphics[width=0.9\textwidth]{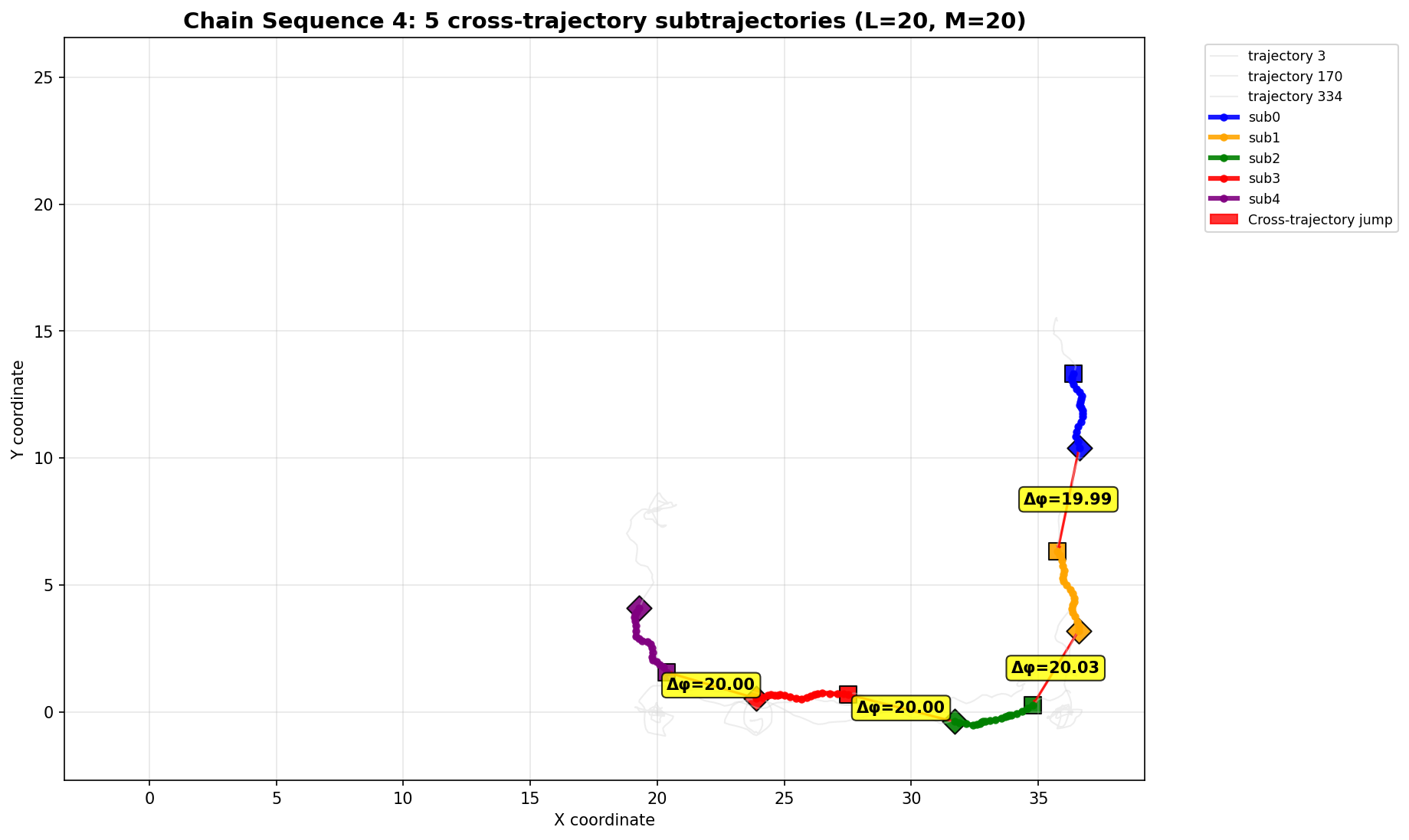}
    \caption{Visualization of selection results 4.}

\end{figure*}

\end{document}